\documentclass{article}

\usepackage{PRIMEarxiv}
\usepackage{amsmath}
\usepackage{amssymb}
\usepackage{bbm} 
\usepackage[ruled,vlined]{algorithm2e}
\usepackage[utf8]{inputenc} 
\usepackage[T1]{fontenc}    
\usepackage{hyperref}       
\usepackage{url}            
\usepackage{booktabs}       
\usepackage{amsfonts}       
\usepackage{nicefrac}       
\usepackage{microtype}      
\usepackage{lipsum}
\usepackage{fancyhdr}       
\usepackage{graphicx}       
\graphicspath{{media/}}     

\title{Neuro-Symbolic Control with Large Language Models
for Language-Guided Spatial Tasks
 
}

\author{
  Momina Liaqat Ali\\
  Department of Computer Science, \\
  Middle Tennessee State University, \\
  Murfreesboro, TN 37130, USA. \\
  \texttt{ma2mp@mtmail.mtsu.edu} \\
  \And
  Muhammad Abid \\
  Department of Mechanical and Aerospace Engineering, \\
  University of Tennessee, \\
  Knoxville, TN 37996, USA. \\
  \texttt{mabid@vols.utk.edu} \\
\And
Muhammad Saqlain \\
  School of Computer Science,\\ Faculty of Engineering, and Information Technology, \\
  University of Technology Sydney, \\
  Broadway, Ultimo, NSW 2007, Australia.\\
\texttt{msgondal0@gmail.com}\\
\And
Jose M. Merigo\thanks{Corresponding author. Email: Jose.Merigo@uts.edu.au} \\
  School of Computer Science, \\ Faculty of Engineering and Information Technology, \\
  University of Technology Sydney, \\
  Broadway, Ultimo, NSW 2007, Australia.\\
\texttt{Jose.Merigo@uts.edu.au}
}

\begin{document}
\maketitle

\begin{abstract}
Although large language models (LLMs) have recently become effective tools for language-conditioned control in embodied systems, instability, slow convergence, and hallucinated actions continue to limit their direct application to continuous control. A modular neuro-symbolic control framework that distinguishes between low-level motion execution and high-level semantic reasoning is proposed in this work. While a lightweight neural delta controller performs bounded, incremental actions in continuous space, a locally deployed LLM interprets symbolic tasks. We assess the suggested method in a planar manipulation setting with spatial relations between objects specified by language. Numerous tasks and local language models, such as Mistral, Phi, and LLaMA-3.2, are used in extensive experiments to compare LLM-only control, neural-only control, and the suggested LLM+Deep Learning (LLM+DL) framework. In comparison to LLM-only baselines, the results show that the neuro-symbolic integration consistently increases both success rate and efficiency, achieving average step reductions exceeding 70\% and speedups of up to $8.83\times$ while remaining robust to language model quality. The suggested framework enhances interpretability, stability, and generalization without any need of reinforcement learning or costly rollouts by controlling the LLM to symbolic outputs and allocating uninterpreted execution to a neural controller trained on artificial geometric data. These outputs show empirically that neuro-symbolic decomposition offers a scalable and principled way to integrate language understanding with ongoing control, this approach promotes the creation of dependable and effective language-guided embodied systems.
\end{abstract}


\section{Introduction}
\label{sec:introduction}

\begin{figure}[ht!]
    \centering
    \includegraphics[width=0.90\linewidth]{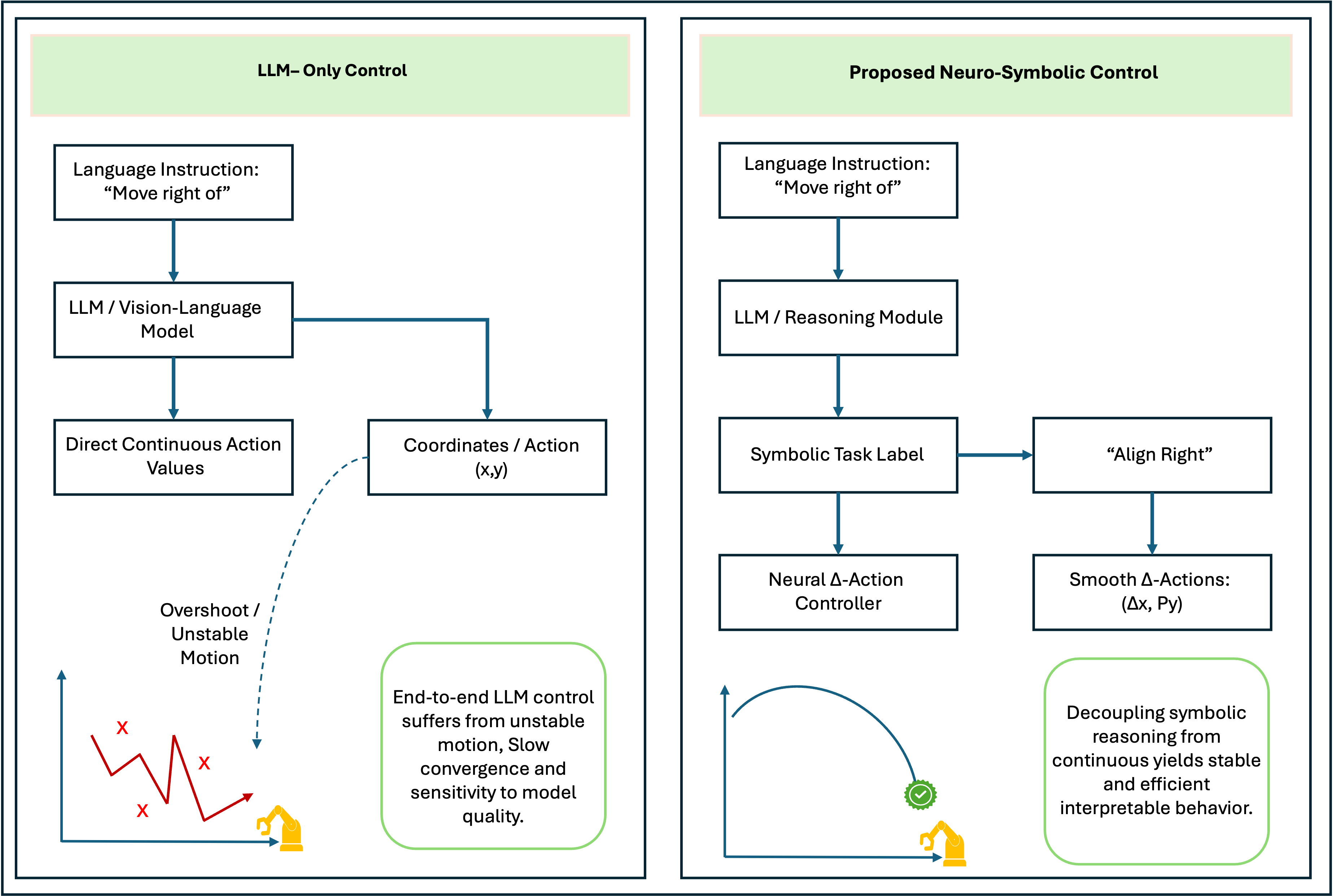}
    \caption{Motivation for neuro-symbolic control.
    End-to-end LLM-based control directly predicts continuous actions,
    leading to unstable motion and slow convergence.
    Our approach decouples symbolic reasoning from continuous execution,
    combining LLM-based semantic understanding with a neural delta controller
    for stable and efficient closed-loop control.}
    \label{fig:intro_motivation}
\end{figure}

Recent developments in large language models (LLMs) have shown their exceptional capacity for high-level decision-making, reasoning, and instruction following \cite{zhang2024llmgraph, bandyopadhyay2025thinking}. These kinds of models are being studied extensively as cognitive engines for robotics and control systems, where they are employed to create the different tasks and plans, choose actions in complex environments, and interpret natural language instructions. However, when used in closed-loop control settings, LLMs show basic limitations despite their expressive power \cite{liu2024optimizing}. Particularly, LLM-only control policies typically suffer from hallucinated actions, poor sample efficiency, lack of convergence guarantees, and brittle behavior under feedback, which makes them unreliable for precise spatial manipulation and continuous control tasks \cite{banerjee2025hallucinate, huang2025hallucination}.
As illustrated in Figure~\ref{fig:intro_motivation}, directly using large
language models for continuous control leads to unstable behavior, motivating
the proposed neuro-symbolic decomposition.

In parallel, learning-based controllers such as deep neural networks have achieved strong performance in low-level control and perception tasks, benefiting from fast inference, smooth action outputs, and stability under feedback. However, these models typically lack symbolic reasoning and struggle to generalize across tasks without retraining \cite{tang2024drl}. Purely neural controllers are therefore limited in their ability to handle task-level abstraction, compositional instructions, or symbolic relational reasoning such as spatial constraints (e.g., \textit{left of}, \textit{right of}, \textit{above}, \textit{below}).

This dichotomy has motivated growing interest in neuro-symbolic control, which seeks to combine the complementary strengths of symbolic reasoning systems and neural function approximators. In this paradigm, symbolic components provide task-level reasoning and interpretability, while neural modules handle continuous control and execution. While prior work has explored hybrid approaches using classical symbolic planners or handcrafted logic modules \cite{du2025fast}, the emergence of LLMs offers a new opportunity: using language models as flexible, general-purpose symbolic planners that can reason over tasks, goals, and constraints expressed in natural language.

Nevertheless, directly deploying LLMs as action generators in control loops still remains problematic \cite{bhat2024grounding}. LLM outputs are fundamentally stochastic, sensitive to prompt phrasing, and also not optimized for numerical precision \cite{su2025statistical}. These type issues become particularly pronounced in spatial control tasks, where the small positional errors can lead to task failure or also towards slow convergence. In term of the result in this scenario, recent studies have highlighted the need for structured interfaces and also the auxiliary control mechanisms to ground LLM reasoning in stable, low-level execution \cite{enoasmo2025structural,ullah2022}.

In the current research work, we propose a neuro-symbolic closed-loop control framework that precisely integrates local large language models with a lightweight type of neural delta controller. So, the LLM is responsible for high-level symbolic reasoning, task interpretation, and also for the goal selection, while the neural controller executes fine-grained corrective actions that ensure fast and stable convergence. By empowering the numerical control to the neural module and restricting the LLM to structured symbolic outputs, the proposed framework mitigates hallucinations, reduces action variance, and significantly improves the sample efficiency.

A fundamental aspect of our proposed approach is the use of locally hosted LLMs, including Mistral, Phi, and LLaMA-3.2. Unlike cloud-based proprietary models, local LLMs facilitate reproducibility, low-latency inference, privacy preservation, and deployment on edge or also resource-constrained systems. Local models, although they are often smaller and less robust than large proprietary counterparts, make them one of the best testbeds for evaluating whether neuro-symbolic integration can compensate for limited model capacity. Our results  demonstrate that the proposed hybrid framework consistently improves performance across all evaluated local LLMs, also indicating that the improvements arise from architectural synergy rather than the model scale alone.

A fundamental aspect of our proposed approach is the use of locally hosted LLMs, including Mistral, Phi, and LLaMA-3.2. Unlike cloud-based proprietary models, local LLMs facilitate reproducibility, low-latency inference, privacy preservation, and deployment on edge or also resource-constrained systems. Local models, although they are often smaller and less robust than large proprietary counterparts, make them one of the best testbeds for evaluating whether neuro-symbolic integration can compensate for limited model capacity. Our results  demonstrate that the proposed hybrid framework consistently improves performance across all evaluated local LLMs, also indicating that the improvements arise from architectural synergy rather than the model scale alone.

We evaluate our method on a suite of spatial reasoning and control tasks in a simulated 2D environment, including relational goals such as \textit{right of}, \textit{left of}, \textit{above}, and \textit{below}. We compare three settings: LLM-only control, deep learning (DL)-only control, and the proposed LLM+DL hybrid. Extensive experiments show that the hybrid approach achieves substantially higher success rates, reduces the number of control steps by up to an order of magnitude, and yields significant speedups over LLM-only baselines. These improvements are consistent across many different types of LLM architectures and tasks, highlighting the robustness and generality of the proposed framework.

Here are the main and clear contributions of this paper as follows:
\begin{itemize}
    \item We introduce a neuro-symbolic closed-loop control architecture that combines LLM-based symbolic reasoning along neural delta controller for stable execution.
    \item We provide a rigorous evaluation of the local LLMs (including Mistral, Phi, and LLaMA-3.2) in spatial control tasks, those highlighting their limitations in isolation and their strengths when integrated with neural control.
    \item We show the major improvements in success rate, convergence speed, and also efficiency using the proposed hybrid framework, supported by a comprehensive quantitative analysis and ablation studies.
    \item We release a reproducible experimental pipeline and evaluation framework that facilitates further research on neuro-symbolic control with local LLMs.
\end{itemize}

\section{Related Work}
\label{sec:relatedwork}

This work intersects several active research areas, including large language models for control, neuro-symbolic reasoning, hybrid planning and control architectures, and learning-based spatial reasoning. We review the most relevant literature and highlight how our approach differs from prior work.

\subsection{Large Language Models for Planning and Control}

Large language models have recently been explored as high-level planners and controllers for robotic systems due to their strong reasoning and instruction-following capabilities~\cite{kim2024survey,zeng2023llmsurvey}. Early work demonstrated that LLMs can generate action sequences, code, or symbolic plans from natural-language task descriptions, enabling flexible task specification and zero-shot generalization~\cite{huang2022language}. Examples include language-driven planning frameworks that translate instructions into executable programs or symbolic action graphs.

Several landmark studies have applied LLMs directly to robotic control loops. SayCan~\cite{ahn2022saycan} demonstrated that combining LLM-based task planning with learned affordance functions enables robots to execute complex, multi-step instructions in real kitchen environments. PaLM-E~\cite{driess2023palme} extended this paradigm by incorporating multimodal sensor inputs directly into the language model, enabling end-to-end embodied reasoning across manipulation and navigation tasks. Code-as-Policies~\cite{liang2023code} showed that code-writing LLMs can generate executable robot policies that exhibit spatial-geometric reasoning and generalize to new instructions without additional training.

While promising, LLM-only approaches often suffer from instability, hallucinated actions, and slow convergence due to the lack of grounding in physical dynamics~\cite{liu2024llmrobotics}. According to empirical investigations, LLM-only controllers perform poorly on tasks those requiring exact numerical reasoning or spatial accuracy~\cite{wang2024survey, mabid2025uq}, show considerable variance between trials, and are sensitive to prompt phrasing. In most recent work, that has attempted to mitigate these issues through prompt engineering, structured outputs (e.g., JSON schemas), or iterative replanning~\cite{chen2024autotamp}. However, such methodologies do not essentially address the mismatch among the discrete language reasoning and continuous control execution. But our work explicitly separates symbolic reasoning and numerical control, using the LLM only for high-level task with interpretation while delegating execution to a neural controller trained for stability and precision.

\subsection{Neuro-Symbolic and Hybrid Reasoning Systems}

Neuro-symbolic AI aims to combine the comprehensibility and also the compositionality of symbolic reasoning with the scalability and robustness of neural networks~\cite{garcez2023neurosymbolic,wan2024cognitive}. Classical methods especially integrate logic-based systems, rule engines, or symbolic planners with neural perception or control modules~\cite{deraedt2024statistical}. These type of methods have been applied in domains such as visual reasoning, program induction, and task planning~\cite{colelough2024neurosymbolic, mabid2024optimizing}.

In the robotics field, hybrid symbolic-neural architectures often employ symbolic task planners (e.g., STRIPS, PDDL-based planners) coupled with some low-level controllers learned via reinforcement learning or also supervised learning~\cite{garrett2021tamp}. While effective, these kinds of approaches typically depend on the handcrafted symbolic representations and domain-specific planners, limiting flexibility and scalability.

In the recent advances, they  propose replacing classical symbolic planners with LLMs, leveraging their ability to reason over diverse tasks without explicit domain modeling~\cite{zhao2024taskplanning}. However, many existing LLM-based neuro-symbolic systems specially focus on open-loop planning or single-shot decision-making rather than closed-loop control~\cite{bousetouane2025agentic}. But our work extends this paradigm by embedding the LLM reasoning within a closed-loop control framework, where symbolic decisions are continuously corrected by a learned neural controller, which is enabling fast convergence and robustness to noise.

\subsection{Learning-Based Controllers and Delta Control}

Learning-based controllers, especially deep neural networks, have demonstrated a strong performance in continuous control tasks due to their ability to approximate the nonlinear dynamics and also to generate smooth action trajectories~\cite{arulkumaran2017deep}. For stabilization and fine-grained adjustments, delta controllers, which forecast incremental state changes rather than absolute actions, are particularly useful~\cite{thanh2020delta,nguyen2021neural}.

Prior work has shown that delta-based control significantly improves convergence speed, reduces overshooting, and also enhances robustness in manipulation and navigation tasks~\cite{zhang2022incremental}. These types of controllers are typically trained using supervised learning or reinforcement learning and also operate efficiently at inference time~\cite{lima2024delta}. Neural network-based trajectory tracking for parallel robots has demonstrated that adaptive learning can also compensate for modeling uncertainties and external disturbances~\cite{mabidmomin2025, khosravi2022adaptive}.

In our proposed technique, the neural delta controller has a crucial role in grounding symbolic decisions which are generated by the LLM. Rather than focusing on replacing the LLM, the controller complements it by making sure that each high-level decision is executed with minimal error. This strict integration especially enables our system to achieve both symbolic generalization and numerical precision. This combination is difficult to achieve with either component alone.

\subsection{Spatial Reasoning and Relational Control Tasks}

Spatial reasoning tasks involving relational concepts, including \textit{left of}, \textit{right of}, \textit{above}, and \textit{below,} have long been used as benchmarks for evaluating reasoning and control systems~\cite{chen2024spatialvlm}. Classical approaches only rely on geometric constraints and optimization-based controllers, while learning-based methods often encode spatial relationships implicitly in the latent representations.

LLMs have shown favorable performance on symbolic spatial reasoning in purely textual domains, but also transferring this capability to physical or simulated control environments remains challenging~\cite{rana2023sayplan}. The prior studies report that especially LLMs struggle with consistent spatial grounding, often producing actions that violate geometric constraints or converge slowly~\cite{hunt2024survey}.

Our current work provides a systematic evaluation of spatial relational control using local LLMs, revealing consistent performance gaps in LLM-only settings. By incorporating a special neural execution module, we demonstrate that spatial reasoning can be reliably converted into precise control behavior, even when using relatively small local language models.

\subsection{Local and Resource-Constrained Large Language Models}

Most of the previous work on LLM-based robotics depends on large proprietary models accessed via cloud APIs~\cite{mabidsama2025,wang2024llmrobotics}. While effective, such approaches raise concerns regarding reproducibility, latency, privacy, and deployment feasibility~\cite{techtalks2024}. Recently, there has been growing interest in local LLMs that can run on consumer-grade hardware, driven by advances in model compression, quantization, and efficient architectures~\cite{wan2024edge}.

Local models such as Mistral~\cite{jiang2023mistral}, Phi~\cite{abdin2024phi}, and LLaMA variants~\cite{touvron2023llama} provide a lower inference costs and also improved controllability but typically exhibit weaker reasoning performance compared to larger models~\cite{datasciencedojo2024}. These such models have been optimized for edge deployment scenarios, also including on-device translation, offline assistants, and autonomous robotics~\cite{mistral2024ministral}. Few studies have also evaluated how  this architectural integration with neural controllers can compensate for these limitations.

In  our proposed work, we directly address this gap by conducting a comparative study across multiple local LLMs and demonstrating that the proposed neuro-symbolic architecture yields consistent gains regardless of model capacity. These findings suggest that architectural design is as significant as model scale in LLM-based control systems.

Previous research has explored LLMs for planning, neuro-symbolic reasoning, and learning-based control in isolation. However, current approaches either depend on open-loop reasoning, handcrafted symbolic planners, or also cloud-scale LLMs. Our work differs by the following points:
\begin{itemize}
    \item Employing the local LLMs as symbolic planners,
    \item Embedding them within a closed-loop neuro-symbolic control architecture, and
    \item Demonstrating quantitative efficiency improvements in the spatial control tasks through tight integration with a neural delta controller.
\end{itemize}
These positions all show our current approach as a practical and scalable solution for grounded, interpretable, and also efficient LLM-based control.


\section{Methodology}
\label{sec:methodology}
In addition to setting the problem and describing the design of each system component, this part provides the specific range of the suggested neuro-symbolic control that is our proposed framework. The methodology differentiates between symbolic reasoning and continuous control while focusing on the modularity, interpretability, and efficiency.

\subsection{Problem Formulation}

We consider a planar manipulation environment containing the following two entities:
\begin{itemize}
    \item A \emph{reference marker} (blue),
    \item A \emph{target marker} (red).
\end{itemize}

The objective is to reposition the target marker such that it satisfies a specified spatial relation regarding the reference marker, as described by a natural language instruction.


Let $\mathcal{W} \subset \mathbb{R}^2$ denote the bounded workspace. At discrete time step $t \in \{0, 1, \ldots, T\}$, the environment state is represented as:
\begin{equation}
\mathbf{s}_t = (x_r^t, y_r^t, x_b^t, y_b^t)^\top \in \mathcal{S} \subseteq \mathbb{R}^4,
\end{equation}
where $(x_r^t, y_r^t)$ denote the Cartesian coordinates of the target marker and $(x_b^t, y_b^t)$ denote those of the reference marker. 

The workspace is bounded by:
\begin{equation}
\mathcal{W} = \{(x, y) \in \mathbb{R}^2 : 0 \leq x \leq C, \; 0 \leq y \leq C\},
\end{equation}
where $C \in \mathbb{R}^+$ is the maximum coordinate value. The state space is thus defined as the Cartesian product $\mathcal{S} = \mathcal{W} \times \mathcal{W}$.
Here, the system state explicitly represents the Cartesian positions of both the target and reference markers within a bounded planar workspace. This formulation provides a minimal yet sufficient representation for spatial relational tasks while avoiding unnecessary state complexity. The bounded workspace constraint ensures that all control actions remain physically valid throughout closed-loop execution.

Each episode is associated with a task $\mathcal{T}$ specified in natural language and mapped to one of four canonical spatial relations:
\begin{equation}
\mathcal{T} \in \Omega = \{\texttt{right\_of}, \texttt{left\_of}, \texttt{above}, \texttt{below}\}.
\end{equation}

A margin parameter $m \in \mathbb{R}^+$ defines tolerance for task satisfaction. We formalize the task satisfaction conditions using the predicate $\mathcal{G}: \mathcal{S} \times \Omega \rightarrow \{0, 1\}$:
\begin{equation}
\mathcal{G}(\mathbf{s}_t, \mathcal{T}) = 
\begin{cases}
\mathbbm{1}[x_r^t \geq x_b^t + m] & \text{if } \mathcal{T} = \texttt{right\_of}, \\[4pt]
\mathbbm{1}[x_r^t \leq x_b^t - m] & \text{if } \mathcal{T} = \texttt{left\_of}, \\[4pt]
\mathbbm{1}[y_r^t \leq y_b^t - m] & \text{if } \mathcal{T} = \texttt{above}, \\[4pt]
\mathbbm{1}[y_r^t \geq y_b^t + m] & \text{if } \mathcal{T} = \texttt{below},
\end{cases}
\end{equation}
where $\mathbbm{1}[\cdot]$ denotes the indicator function returning 1 if the condition holds and 0 otherwise. The task space is restricted to a small set of canonical spatial relations, allowing the symbolic reasoning module to operate over a discrete and interpretable output space. The goal predicate $G(\cdot)$ defines task satisfaction using margin-based geometric constraints, which introduces tolerance against small positional errors and supports stable convergence near task boundaries.

The action space consists of incremental displacements bounded by maximum step size $\delta_{\max}$:
\begin{equation}
\mathcal{A} = \{(\Delta x, \Delta y) \in \mathbb{R}^2 : |\Delta x| \leq \delta_{\max}, \; |\Delta y| \leq \delta_{\max}\}.
\end{equation}


Given an initial state $\mathbf{s}_0 \sim \mathcal{P}_0$ and task $\mathcal{T} \sim \mathcal{P}_\mathcal{T}$, the goal is to compute a control policy $\pi: \mathcal{S} \times \Omega \rightarrow \mathcal{A}$ that:
\begin{enumerate}
    \item Achieves task satisfaction within a finite horizon $T$:
    \begin{equation}
    \exists \, t^\ast \leq T : \mathcal{G}(\mathbf{s}_{t^\ast}, \mathcal{T}) = 1,
    \end{equation}
    \item The expected number of control steps minimizes as follows:
    \begin{equation}
    \pi^\ast = \arg\min_{\pi} \; \mathbb{E}_{\mathbf{s}_0, \mathcal{T}} \left[ \sum_{t=0}^{T-1} \left(1 - \mathcal{G}(\mathbf{s}_t, \mathcal{T})\right) \right],
    \end{equation}
    \item Is robust to variations in language models and initial configurations.
    
    The action space is defined in terms of bounded incremental displacements rather than absolute position commands. This choice prevents large, destabilizing movements and encourages smooth trajectories. The optimization objective prioritizes early task satisfaction, thereby directly minimizing control effort and convergence time in expectation.
\end{enumerate}

\subsection{Overall Neuro-Symbolic Architecture}

The proposed framework decomposes decision-making into two interacting layers:
\begin{itemize}
    \item \textbf{Symbolic reasoning layer} $\pi_{\text{sym}}: \mathcal{S} \times \Omega \rightarrow \mathcal{Z}$: responsible for task interpretation and semantic guidance.
    \item \textbf{Neural execution layer} $\pi_{\text{neu}}: \mathcal{S} \times \mathcal{Z} \rightarrow \mathcal{A}$: responsible for continuous control and motion refinement.
\end{itemize}

The composite policy is expressed as:
\begin{equation}
\pi(\mathbf{s}_t, \mathcal{T}) = \pi_{\text{neu}}\left(\mathbf{s}_t, \pi_{\text{sym}}(\mathbf{s}_t, \mathcal{T})\right),
\end{equation}
where the symbolic latent space encoding task semantics denotes by $\mathcal{Z}$. Equation~(8) formalizes the neuro-symbolic decomposition central to the proposed framework. 
High-level symbolic reasoning is handled independently by $\pi_{\mathrm{sym}}$, while continuous 
execution is delegated to the neural controller $\pi_{\mathrm{neu}}$. 
This explicit separation allows semantic interpretation and numerical control to be managed by 
specialized modules, thereby improving control stability, interpretability, and overall robustness.

This separation follows the principle that symbolic abstractions and also the continuous dynamics should be managed by specialized modules rather than a monolithic policy.

\subsection{Symbolic Reasoning Layer}

The symbolic layer implements a locally deployed large language model (LLM) $\mathcal{M}_{\text{LLM}}$ to process the natural language instruction and also reason about the desired spatial relationship.


The LLM receives a structured prompt at each time step $t$ as follows:
\begin{equation}
\mathbf{q}_t = \langle \mathcal{T}_{\text{nl}}, \psi(\mathbf{s}_t) \rangle,
\end{equation}
where the natural language task description is denoted as $\mathcal{T}_{\text{nl}}$ and $\psi: \mathcal{S} \rightarrow \Sigma^\ast$ is a serialization function mapping states to structured text over alphabet $\Sigma$. 

To ensure the validity of the LLM that is constrained to output strictly formatted JSON responses as follows:
\begin{equation}
\mathbf{z}_t = \text{Parse}\left(\mathcal{M}_{\text{LLM}}(\mathbf{q}_t)\right) \in \mathcal{Z},
\end{equation}
where $\mathcal{Z} = \{0, 1, 2, 3\}$ corresponds to the four canonical spatial relations. The LLM receives a structured prompt combining the natural language task description with a serialized representation of the current state. Importantly, the LLM output is constrained to a fixed symbolic space through structured parsing. This restriction prevents hallucinated continuous actions and ensures that the language model influences control only through interpretable, discrete decisions.

The LLM does not produce continuous actions directly. This constraint prevents instability, hallucinated coordinates, and non-physical behaviors commonly observed in LLM-only control.


The framework supports interchangeable local LLMs:
\begin{equation}
\mathcal{M}_{\text{LLM}} \in \{\text{Mistral}, \text{Phi}, \text{LLaMA-3.2}\},
\end{equation}
without retraining the execution policy. This design enables systematic evaluation of language reasoning quality while keeping the control layer fixed. By allowing interchangeable local language models without retraining the neural controller, the framework enables a controlled evaluation of symbolic reasoning quality. This design isolates the effect of the language model from low-level execution dynamics, making performance differences directly attributable to semantic inference rather than control capacity.

\subsection{Neural Delta Controller}

The execution layer is implemented as a feedforward neural network $f_\theta: \mathbb{R}^d \rightarrow \mathbb{R}^2$ with parameters $\theta$, trained to predict incremental displacements toward task satisfaction.

\subsubsection{Input Encoding}

The controller input at time $t$ is constructed by concatenating normalized state features with the task encoding:
\begin{equation}
\mathbf{u}_t = \left[ \frac{x_r^t}{C}, \; \frac{y_r^t}{C}, \; \frac{x_b^t}{C}, \; \frac{y_b^t}{C}, \; \phi(\mathcal{T}) \right]^\top \in \mathbb{R}^5,
\end{equation}
where $\phi: \Omega \rightarrow [0, 1]$ is a scalar encoding of the task relation defined as:
\begin{equation}
\phi(\mathcal{T}) = \frac{\text{idx}(\mathcal{T})}{|\Omega| - 1},
\end{equation}
with $\text{idx}(\cdot)$ returning the index of the task in $\Omega$.

Alternatively, a one-hot encoding $\phi_{\text{oh}}: \Omega \rightarrow \{0,1\}^{|\Omega|}$ may be used:
\begin{equation}
\phi_{\text{oh}}(\mathcal{T}) = \mathbf{e}_{\text{idx}(\mathcal{T})} \in \{0,1\}^4,
\end{equation}
where $\mathbf{e}_i$ denotes the $i$-th standard basis vector, yielding input dimension $d = 8$. The neural controller input combines normalized state features with a compact encoding of the symbolic task label. Normalization ensures numerical stability during training and inference, while the task encoding provides explicit semantic conditioning. Both scalar and one-hot encodings are supported, allowing flexibility without altering the controller architecture.


The neural controller consists of $L$ fully-connected layers:
\begin{equation}
f_\theta(\mathbf{u}) = \mathbf{W}_L \, \sigma\left(\mathbf{W}_{L-1} \, \sigma\left(\cdots \sigma\left(\mathbf{W}_1 \mathbf{u} + \mathbf{b}_1\right) \cdots\right) + \mathbf{b}_{L-1}\right) + \mathbf{b}_L,
\end{equation}
where $\sigma(\cdot) = \max(0, \cdot)$ is the ReLU activation, $\mathbf{W}_l \in \mathbb{R}^{n_l \times n_{l-1}}$ are weight matrices, and $\mathbf{b}_l \in \mathbb{R}^{n_l}$ are bias vectors. The neural execution module is implemented as a feedforward network that maps state-task representations to incremental control actions. ReLU activations enable efficient learning of nonlinear control policies, while the fully connected structure ensures low inference latency suitable for closed-loop operation.

The output layer applies hyperbolic tangent activation to bound displacements:
\begin{equation}
(\Delta x_t, \Delta y_t) = \delta_{\max} \cdot \tanh\left(f_\theta(\mathbf{u}_t)\right) \in [-\delta_{\max}, \delta_{\max}]^2.
\end{equation}

The hyperbolic tangent activation enforces strict bounds on the predicted displacements, preventing excessive control actions. This bounded delta formulation reduces overshooting and oscillations commonly observed in LLM-only control, leading to smoother trajectories and more stable convergence.


The predicted displacement updates the target marker position via the transition function $\mathcal{F}: \mathcal{S} \times \mathcal{A} \rightarrow \mathcal{S}$:
\begin{align}
x_r^{t+1} &= \text{clip}(x_r^t + \Delta x_t, 0, C), \\
y_r^{t+1} &= \text{clip}(y_r^t + \Delta y_t, 0, C),
\end{align}
where the clipping function enforces workspace boundaries:
\begin{equation}
\text{clip}(z, a, b) = \min(\max(z, a), b).
\end{equation}
The state transition updates only the target marker position while enforcing workspace constraints through clipping. This ensures that the system remains within valid spatial limits throughout execution. Combined with bounded actions, this transition model contributes to safe and predictable closed-loop behavior.

\begin{figure}[ht!]
    \centering
    \includegraphics[width=0.90\linewidth]{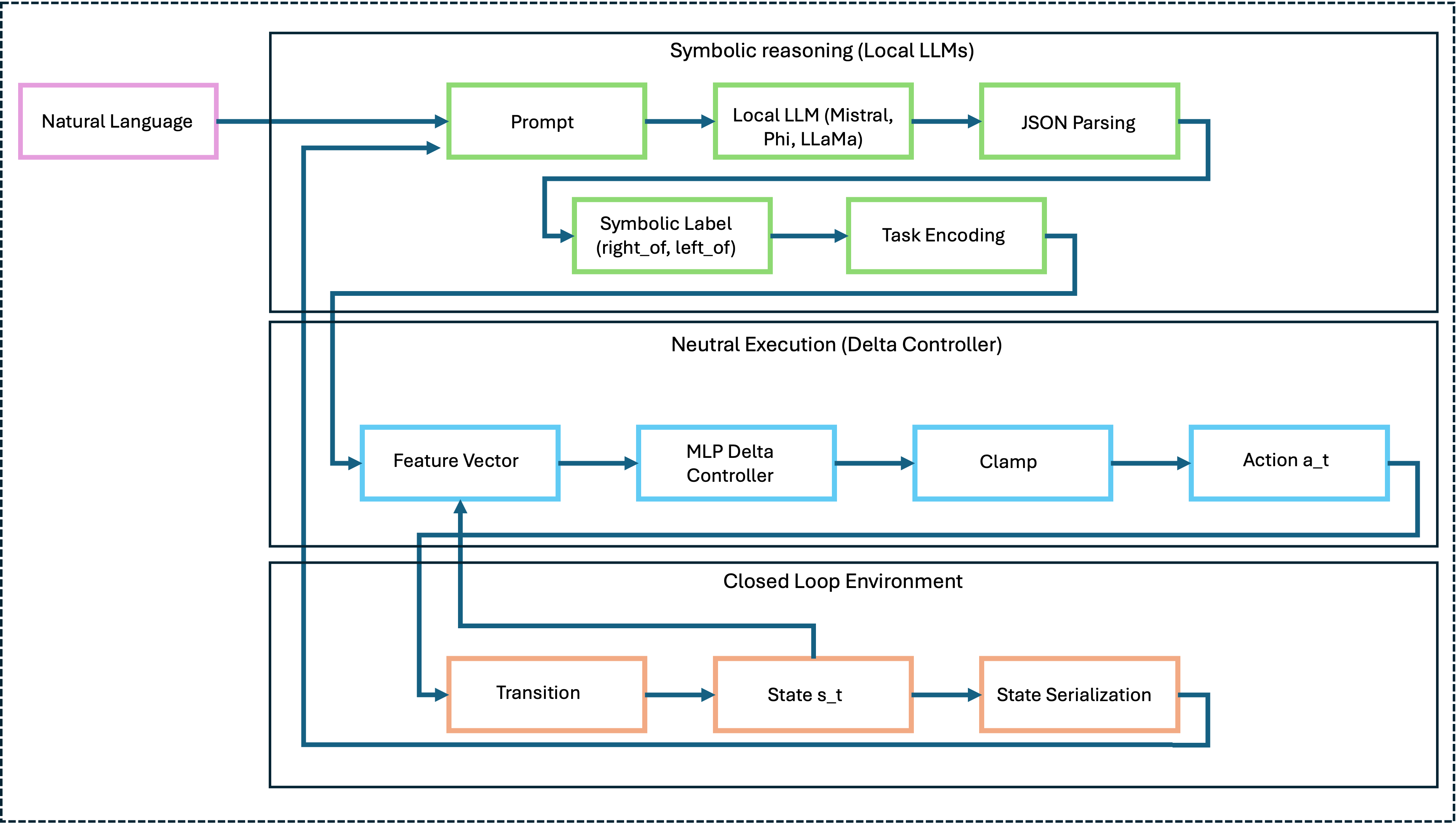}
    \caption{Overview of the proposed neuro-symbolic control framework.
    A local large language model performs symbolic reasoning over language
    instructions and environment state, producing a discrete task label.
    This symbolic output conditions a neural delta controller that executes
    bounded continuous actions in a closed-loop environment, enabling
    stable, efficient, and interpretable control.}
    \label{fig:architecture}
\end{figure}

Delta-based control reduces overshooting and improves stability compared to absolute coordinate prediction. As shown in Figure~\ref{fig:architecture}, the proposed framework
decouples symbolic reasoning from continuous control by delegating language understanding to a local LLM and motion execution to a neural delta controller.

\subsection{Training Procedure}

The neural controller is trained using supervised learning on synthetically generated trajectories.


Each training sample consists of a state-task-displacement tuple:
\begin{equation}
\mathcal{D} = \left\{ \left(\mathbf{s}^{(i)}, \mathcal{T}^{(i)}, \Delta\mathbf{p}^{\ast(i)}\right) \right\}_{i=1}^{N},
\end{equation}
where $\mathbf{s}^{(i)} \sim \text{Uniform}(\mathcal{S})$, $\mathcal{T}^{(i)} \sim \text{Uniform}(\Omega)$, and $\Delta\mathbf{p}^{\ast(i)} = (\Delta x^\ast, \Delta y^\ast)$ is the target displacement that moves the target marker closer to task satisfaction.


The optimal displacement is computed as the direction toward the goal region scaled by step size $\alpha \in (0, 1]$. For $\mathcal{T} = \texttt{right\_of}$:
\begin{equation}
\Delta x^\ast = \alpha \cdot \max(0, (x_b + m) - x_r), \quad \Delta y^\ast = 0.
\end{equation}

Analogous expressions hold for other spatial relations:
\begin{align}
\texttt{left\_of}: & \quad \Delta x^\ast = -\alpha \cdot \max(0, x_r - (x_b - m)), \quad \Delta y^\ast = 0, \\
\texttt{above}: & \quad \Delta x^\ast = 0, \quad \Delta y^\ast = -\alpha \cdot \max(0, y_r - (y_b - m)), \\
\texttt{below}: & \quad \Delta x^\ast = 0, \quad \Delta y^\ast = \alpha \cdot \max(0, (y_b + m) - y_r).
\end{align}


The mean squared error between the target and forecast displacements is the definition of the loss function:
\begin{equation}
\mathcal{L}(\theta) = \frac{1}{N} \sum_{i=1}^{N} \left\| f_\theta(\mathbf{u}^{(i)}) - \Delta\mathbf{p}^{\ast(i)} \right\|_2^2.
\end{equation}

To encourage smooth trajectories, an optional regularization term penalizes large displacement magnitudes:
\begin{equation}
\mathcal{L}_{\text{reg}}(\theta) = \mathcal{L}(\theta) + \lambda \cdot \frac{1}{N} \sum_{i=1}^{N} \left\| f_\theta(\mathbf{u}^{(i)}) \right\|_2^2,
\end{equation}
where $\lambda \geq 0$ is the regularization coefficient. In the algorithm, the closed-loop interaction between symbolic reasoning and neural execution. At each control step, symbolic guidance is recomputed based on the current state, enabling online correction of reasoning errors. This feedback structure allows the system to recover from imperfect symbolic decisions while maintaining efficient convergence.


Parameters $\theta$ are optimized using stochastic gradient descent with the Adam optimizer:
\begin{equation}
\theta_{k+1} = \theta_k - \eta \cdot \frac{\hat{\mathbf{m}}_k}{\sqrt{\hat{\mathbf{v}}_k} + \epsilon},
\end{equation}
where the learning rate is denoted by $\eta$, $\hat{\mathbf{m}}_k$, $\hat{\mathbf{v}}_k$ are bias-corrected moment estimates, and a small constant for numerical stability is denoted by $\epsilon$.

This formulation advocates a smooth, task-directed motion and more generalization across tasks.

\subsection{Closed-Loop Control Strategy}

The following process governs the closed-loop operation of the system is shown in the Algorithm \ref{alg:control} below:

\begin{algorithm}[ht!]
\caption{Neuro-Symbolic Closed-Loop Control}
\label{alg:control}

\KwIn{Initial state $\mathbf{s}_0$, task $\mathcal{T}$, horizon $T$, LLM $\mathcal{M}_{\text{LLM}}$, controller $f_\theta$}
\KwOut{Final state $\mathbf{s}_{t^\ast}$, success flag, step count}

$t \gets 0$\;

\While{$t < T$ \textbf{and} $\mathcal{G}(\mathbf{s}_t, \mathcal{T}) = 0$}{
    Construct prompt $\mathbf{q}_t \gets \langle \mathcal{T}_{\text{nl}}, \psi(\mathbf{s}_t) \rangle$\;
    Symbolic reasoning $\mathbf{z}_t \gets \text{Parse}(\mathcal{M}_{\text{LLM}}(\mathbf{q}_t))$\;
    Encode input $\mathbf{u}_t \gets [\mathbf{s}_t / C, \phi(\mathbf{z}_t)]^\top$\;
    Compute action $(\Delta x_t, \Delta y_t) \gets f_\theta(\mathbf{u}_t)$\;
    Update state $\mathbf{s}_{t+1} \gets \mathcal{F}(\mathbf{s}_t, (\Delta x_t, \Delta y_t))$\;
    $t \gets t + 1$\;
}

\Return $\mathbf{s}_t$, $\mathcal{G}(\mathbf{s}_t, \mathcal{T})$, $t$\;
\end{algorithm}

This feedback loop enables online correction and robust convergence even under imperfect symbolic reasoning.

\subsection{Distance-to-Goal Metric}

To analyze convergence behavior, we define a distance-to-goal metric $d: \mathcal{S} \times \Omega \rightarrow \mathbb{R}_{\geq 0}$:
\begin{equation}
d(\mathbf{s}_t, \mathcal{T}) = 
\begin{cases}
\max(0, (x_b^t + m) - x_r^t) & \text{if } \mathcal{T} = \texttt{right\_of}, \\[4pt]
\max(0, x_r^t - (x_b^t - m)) & \text{if } \mathcal{T} = \texttt{left\_of}, \\[4pt]
\max(0, y_r^t - (y_b^t - m)) & \text{if } \mathcal{T} = \texttt{above}, \\[4pt]
\max(0, (y_b^t + m) - y_r^t) & \text{if } \mathcal{T} = \texttt{below}.
\end{cases}
\end{equation}

The following characteristics are satisfied by this metric:
\begin{enumerate}
    \item \textbf{Non-negativity}: $d(\mathbf{s}_t, \mathcal{T}) \geq 0$ for all $\mathbf{s}_t \in \mathcal{S}$, $\mathcal{T} \in \Omega$.
    \item \textbf{Goal identification}: $d(\mathbf{s}_t, \mathcal{T}) = 0 \Leftrightarrow \mathcal{G}(\mathbf{s}_t, \mathcal{T}) = 1$.
    \item \textbf{Continuity}: $d$ is continuous in $\mathbf{s}_t$.
\end{enumerate}

For cross-episode comparison, we define the normalized distance:
\begin{equation}
\bar{d}_t = \frac{d(\mathbf{s}_t, \mathcal{T})}{d(\mathbf{s}_0, \mathcal{T}) + \epsilon},
\end{equation}
where $\epsilon > 0$ ensures numerical stability when $d(\mathbf{s}_0, \mathcal{T}) \approx 0$. The distance-to-goal metric provides a continuous measure of task progress that is aligned with the discrete success predicate. Normalization by the initial distance enables fair comparison across episodes with different starting configurations. This metric is used to analyze convergence speed and stability across control strategies.

\section{Experimental Setup and Results}
\label{sec:experiments}
This section demonstrated the details of the experimental evaluation of the proposed neuro-symbolic control framework. We establish the baseline processes, evaluation metrics, experimental design, and quantitative results. We designed the experiments in such a way that they ensure reproducibility, controlled ablation, and statistically significant comparisons between control strategies and language models.

\subsection{Environment Configuration}


All experiments are conducted in a planar simulated environment containing two point
markers: a movable target marker (red) and a static reference marker (blue).
The workspace is defined as a bounded square region
$\mathcal{W} = [0, C]^2 \subset \mathbb{R}^2$, with $C = 800$ pixels.

At the beginning of each episode, the positions of the target and reference markers
are independently sampled from a uniform distribution over the workspace.
Episodes are executed for a fixed control horizon $T$, and the episode terminates
early if the task satisfaction condition is met.
All experiments use identical workspace bounds and initialization procedures to
ensure consistent evaluation across methods.

\subsubsection{Task Specification}

We evaluate four canonical spatial relations:
$\{\texttt{right\_of}, \texttt{left\_of}, \texttt{above}, \texttt{below}\}$.
A task is considered successful if the corresponding spatial constraint is satisfied
within a tolerance margin of $m = 50$ pixels at any time step $t \leq T$.

Each task is provided to the system in natural language form.
For example, the instruction \emph{“Move the red object to the right of the blue object”}
corresponds to the \texttt{right\_of} relation.
The same task set and margin values are used across all compared control strategies.

\subsubsection{Compared Methods}

We compare three control strategies to isolate the contribution of symbolic reasoning
and neural execution.

\paragraph{LLM-Only Control.}
In this baseline, a large language model directly predicts absolute target coordinates
at each control step based on the current state and the natural language instruction.
The resulting displacement is computed as the difference between the predicted and
current target positions.
No learned motion prior or bounded control mechanism is applied.

\paragraph{DL-Only Control.}
This baseline uses the neural delta controller exclusively, conditioned on a fixed
oracle task encoding corresponding to the ground-truth spatial relation.
No language model inference is performed during execution.
This configuration serves as an upper bound on execution performance when task
semantics are perfectly specified.

\paragraph{LLM+DL (Proposed).}
The proposed method combines symbolic reasoning and neural execution.
At each control step, a local language model infers a symbolic task label from the
natural language instruction and current state.
This symbolic output conditions the neural delta controller, which generates bounded
incremental displacements in a closed-loop manner.

All methods share the same environment, control horizon, and evaluation protocol.

\subsubsection{Large Language Models}

Symbolic reasoning is performed using locally deployed large language models to
ensure reproducibility, low-latency inference, and independence from external APIs.
The following models are evaluated:

\begin{itemize}
    \item \textbf{Mistral-7B}, a 7B-parameter instruction-tuned model,
    \item \textbf{Phi-2}, a 2.7B-parameter compact model optimized for efficiency,
    \item \textbf{LLaMA-3.2}, an open-weight model with strong general reasoning capability.
\end{itemize}

The neural delta controller $f_\theta$ is kept fixed across all experiments.
This design isolates the effect of symbolic reasoning quality and ensures that
performance differences arise from language model inference rather than changes
in low-level control dynamics.

\subsubsection{Evaluation Metrics}

For each combination of our proposed method, language model, and also the task, we conduct a comprehensive $N = 20$ independent episode with the randomized initial configurations to show the strong results. Also, we showed the mean performance with standard deviation to quantify variability of the proposed framework.


We assess performance using the metrics like success rate \cite{kress2024robot}, average steps and distance to goal \cite{faigl2012goal}:

\paragraph{Success Rate (SR)}
The fraction of episodes achieving task satisfaction within the time horizon:
\begin{equation}
\text{SR} = \frac{1}{N} \sum_{i=1}^{N} \mathbbm{1}\left[\exists \, t \leq T : \mathcal{G}(\mathbf{s}_t^{(i)}, \mathcal{T}) = 1\right].
\end{equation}

\paragraph{Average Steps (AS)}
The mean number of control steps required for successful episodes:
\begin{equation}
\text{AS} = \frac{1}{|\mathcal{S}_{\text{succ}}|} \sum_{i \in \mathcal{S}_{\text{succ}}} t_i^\ast,
\end{equation}
where $\mathcal{S}_{\text{succ}}$ denotes the set of successful episodes and $t_i^\ast$ is the termination step.

\paragraph{Normalized Distance-to-Goal}
The task-specific geometric distance to the satisfaction boundary, normalized by initial distance:
\begin{equation}
\bar{d}_t = \frac{d(\mathbf{s}_t, \mathcal{T})}{d(\mathbf{s}_0, \mathcal{T}) + \epsilon}.
\end{equation}

\paragraph{Relative Improvement Metrics}
To quantify the benefit of neuro-symbolic integration, we compute:
\begin{itemize}
    \item \textbf{Success Rate Improvement}: $\Delta\text{SR} = \text{SR}_{\text{LLM+DL}} - \text{SR}_{\text{LLM-only}}$
    \item \textbf{Step Reduction}: $\rho = \frac{\text{AS}_{\text{LLM-only}} - \text{AS}_{\text{LLM+DL}}}{\text{AS}_{\text{LLM-only}}} \times 100\%$
    \item \textbf{Speedup Factor}: $\sigma = \frac{\text{AS}_{\text{LLM-only}}}{\text{AS}_{\text{LLM+DL}}}$
\end{itemize}

\subsection{Results}
\label{sec:results} 

\begin{figure}[ht!]
    \centering
    \includegraphics[width=0.90\linewidth]{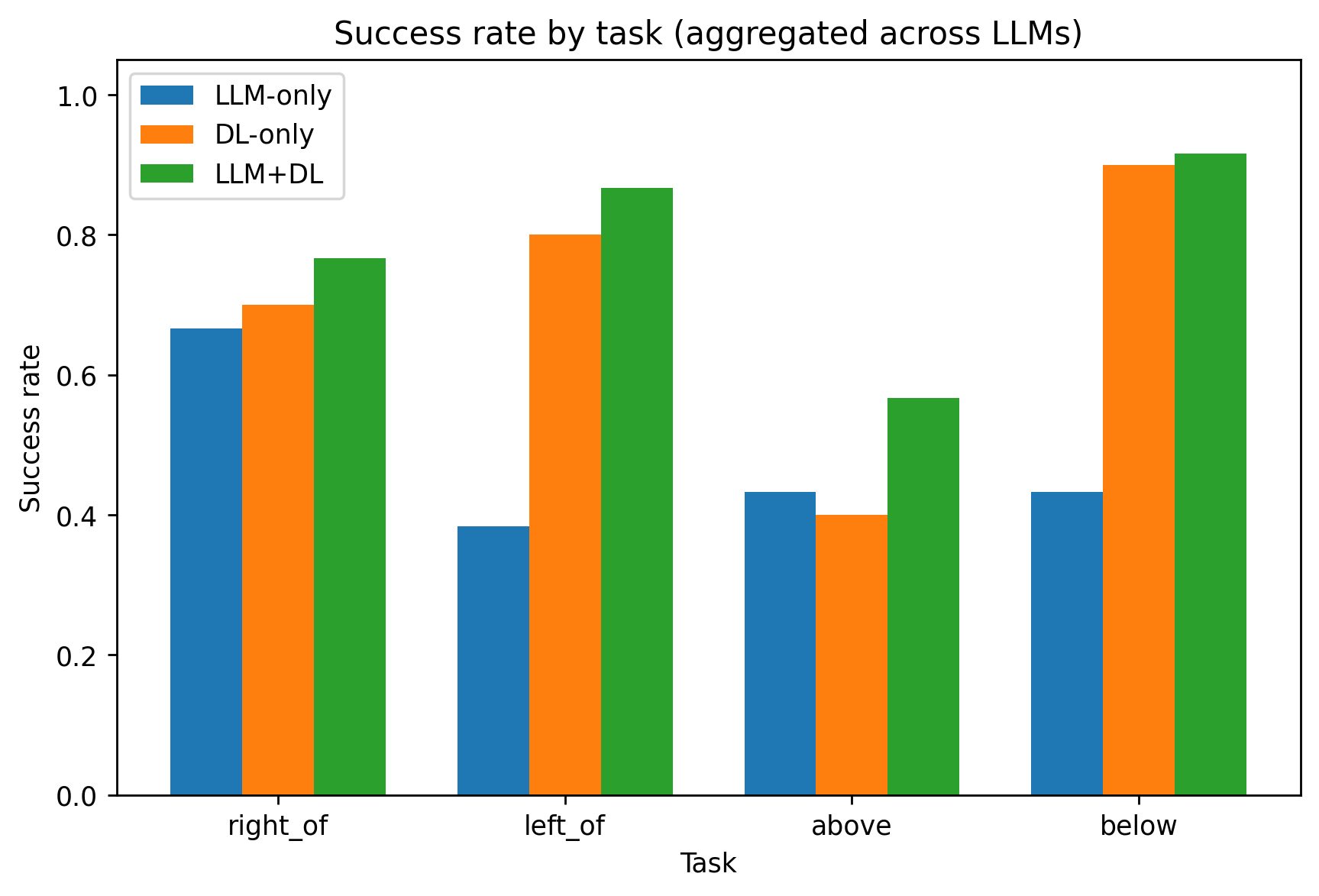}
    \caption{Success rate aggregated across all language models for each spatial task. 
    The proposed LLM+DL framework consistently outperforms LLM-only and DL-only baselines, 
    demonstrating the effectiveness of neuro-symbolic integration.}
    \label{fig:success_by_task_agg}
\end{figure}

In this section, we discussed the results obtained on different LLM models and what are the key improvements and observations emerged from these experiments. In Figure \ref{fig:success_by_task_agg}, we can see that the proposed LLM+DL framework consistently achieves higher success rates across all the spatial tasks when it is compared to both LLM-only and DL-only baselines frameworks.

The average number of steps and success rate for each technique, task, and language model for the \texttt{right\_of} task are summarized in Table~\ref{tab:main_results}. The supplemental material contains the full results for each task.

\begin{table}[t]
\centering
\caption{Performance comparison across control methods and LLMs.}
\label{tab:main_results}
\begin{tabular}{l l l c c}
\hline
LLM Model & Method & Task & Success Rate & Avg. Steps \\
\hline
-- & DL-only & right\_of & 0.70 & 1.35 \\
Mistral & LLM-only & right\_of & 0.70 & 3.60 \\
Mistral & LLM+DL & right\_of & 0.70 & 1.15 \\
Phi & LLM-only & right\_of & 0.60 & 4.95 \\
Phi & LLM+DL & right\_of & 0.85 & 0.85 \\
LLaMA-3.2 & LLM-only & right\_of & 0.70 & 3.40 \\
LLaMA-3.2 & LLM+DL & right\_of & 0.75 & 1.35 \\
\hline
\end{tabular}
\end{table}

Several key observations are emerged from the results. Among which, the first one is, for the consistent improvement, across all language models, the suggested LLM+DL framework produces success rates that are either greater or equivalent to those of LLM-only control, with significant drops in average steps. Secondly, for the efficiency gains, depending on the underlying language model, the neural delta controller allows convergence in many fewer steps, with speedups ranging from $2.52\times$ to $5.82\times$. The third most significant observation was for the compensation for weak reasoning for which the Phi model shows the best LLM+DL success rate (0.85) but the lowest LLM-only success rate (0.60), indicating that the neural controller successfully makes up for weaker symbolic reasoning.


According to Figure \ref{fig:avg_steps_agg}, LLM+DL typically reduces the number of necessary control steps by more than 70\%. Each method's unique qualities are shown by the convergence behavior:

\begin{figure}[ht!]
    \centering
    \includegraphics[width=0.90\linewidth]{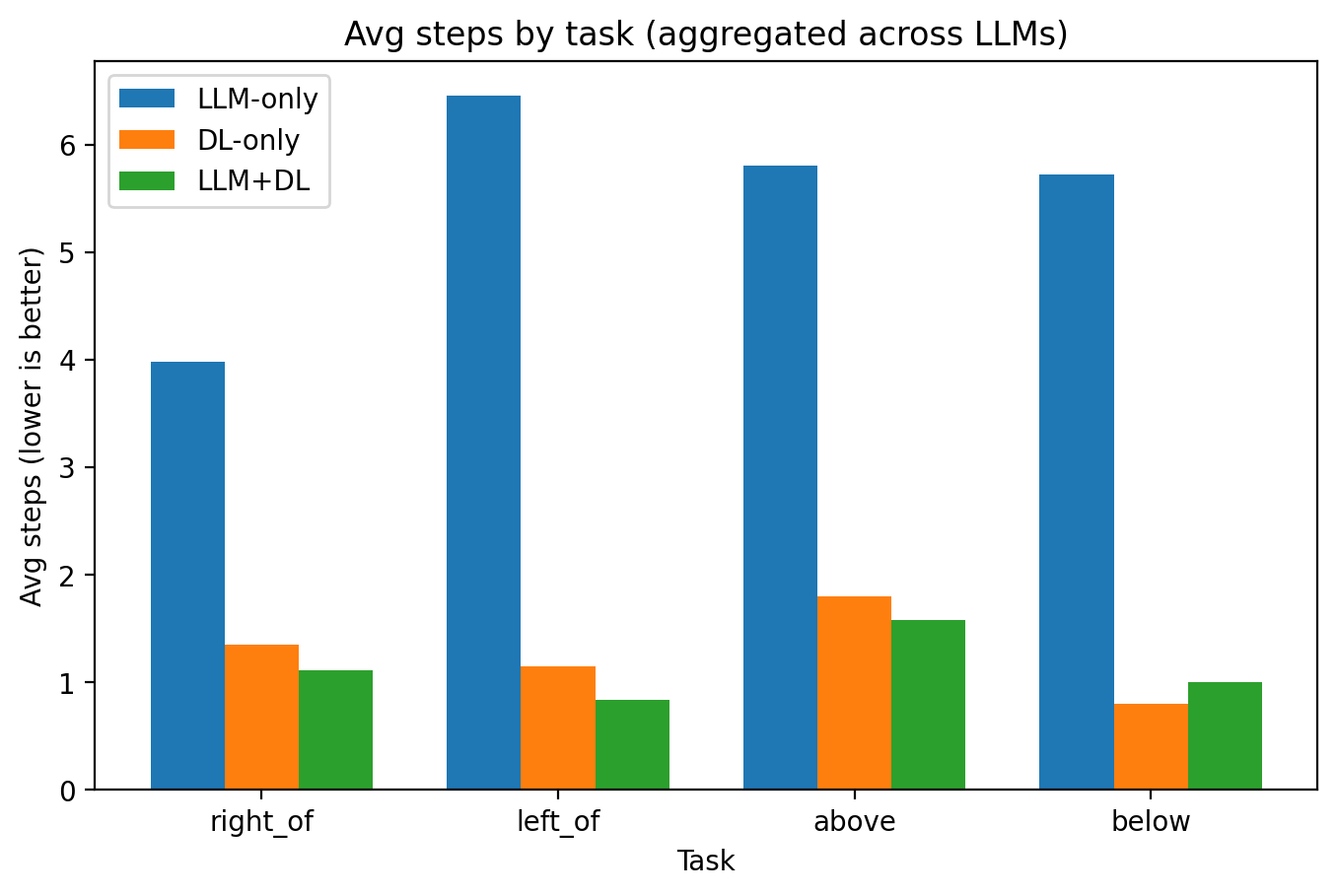}
    \caption{Total average number of control steps for all language models. Compared to LLM-only control, the LLM+DL framework converges far more quickly, resulting in a step reduction of more than 70\% for all jobs.}
    \label{fig:avg_steps_agg}
\end{figure}

\begin{itemize}
    \item For the LLM-only, it shows oscillatory behavior as a result of inaccurate coordinate predictions and sluggish convergence with large variance. The distance-to-goal curve exhibits numerous reversals and non-monotonic descent.
    
    \item For the DL-only, when given Oracle task encoding, it converges quickly and monotonically. Nevertheless, new task descriptions or variations in natural language cannot be accommodated by this setup.
    
    \item For the LLM+DL, it retains the capacity to understand normal language instructions while achieving quick, consistent convergence on par with DL-only. While symbolic reasoning offers high-level semantic guidance, the neural delta controller guarantees smooth paths.
\end{itemize}

\begin{figure}[ht!]
    \centering
    \includegraphics[width=0.90\linewidth]{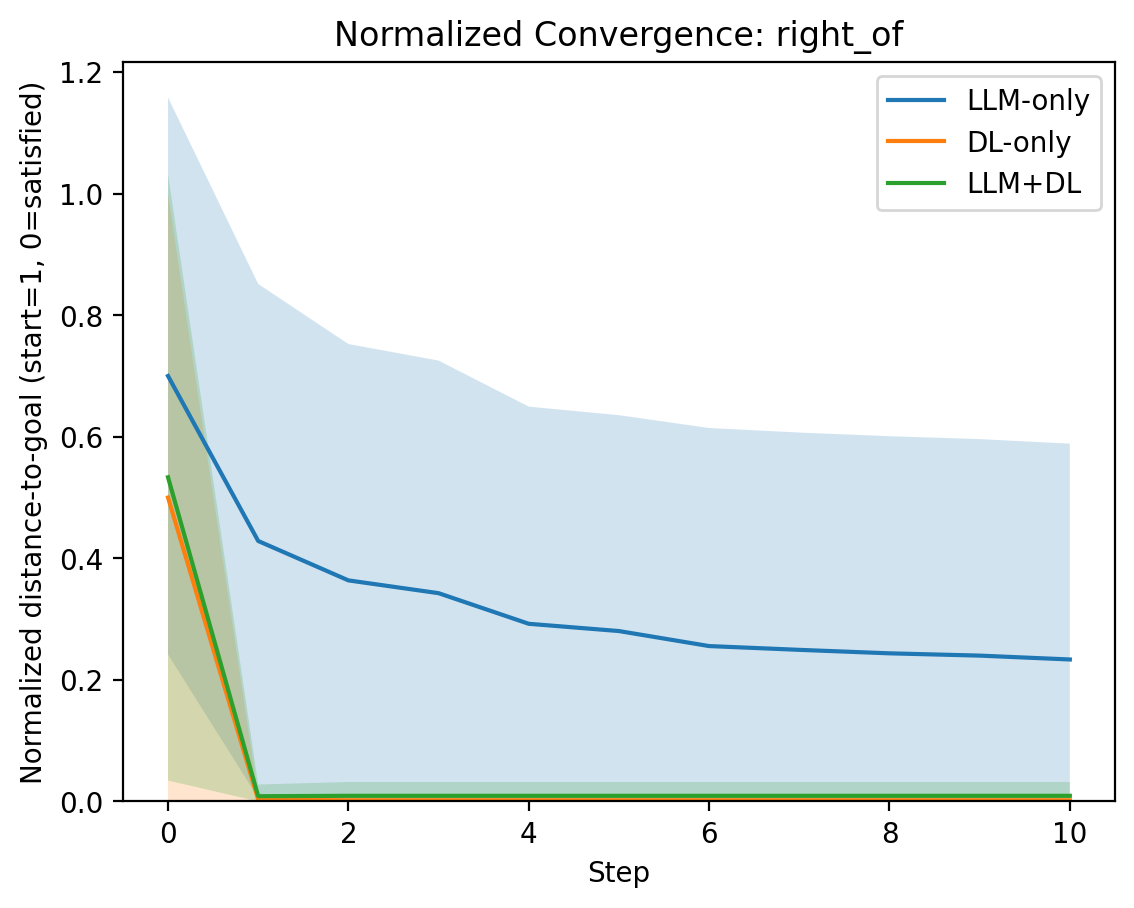}
    \caption{Normalized distance-to-goal over time for the \texttt{right\_of} task. 
    LLM+DL exhibits fast, monotonic convergence with low variance, whereas the LLM-only 
    control shows slower and less stable behavior.}
    \label{fig:convergence_norm}
\end{figure}

The normalized distance-to-goal over time is shown in Figure \ref{fig:convergence_norm}, which highlights the LLM+DL framework's faster and more stable convergence.

To quantify the benefit of neuro-symbolic integration across all tasks, Table~\ref{tab:relative_improvement} reports aggregated statistics comparing LLM+DL to LLM-only control. Results are averaged across all three language models.

\begin{table}[t]
\centering
\caption{Relative improvement of LLM+DL over LLM-only (mean $\pm$ std).}
\label{tab:relative_improvement}
\begin{tabular}{l c c c}
\hline
Task & $\Delta$ Success & Step Reduction (\%) & Speedup ($\times$) \\
\hline
right\_of & 0.10 $\pm$ 0.13 & 70.4 $\pm$ 11.4 & 3.82 $\pm$ 1.76 \\
left\_of & 0.48 $\pm$ 0.23 & 85.9 $\pm$ 7.7 & 8.83 $\pm$ 4.93 \\
above & 0.13 $\pm$ 0.25 & 72.4 $\pm$ 9.2 & 3.87 $\pm$ 1.09 \\
below & 0.48 $\pm$ 0.10 & 82.4 $\pm$ 3.3 & 5.81 $\pm$ 1.07 \\
\hline
\end{tabular}
\end{table}

\begin{figure}[ht!]
    \centering
    \includegraphics[width=0.90\linewidth]{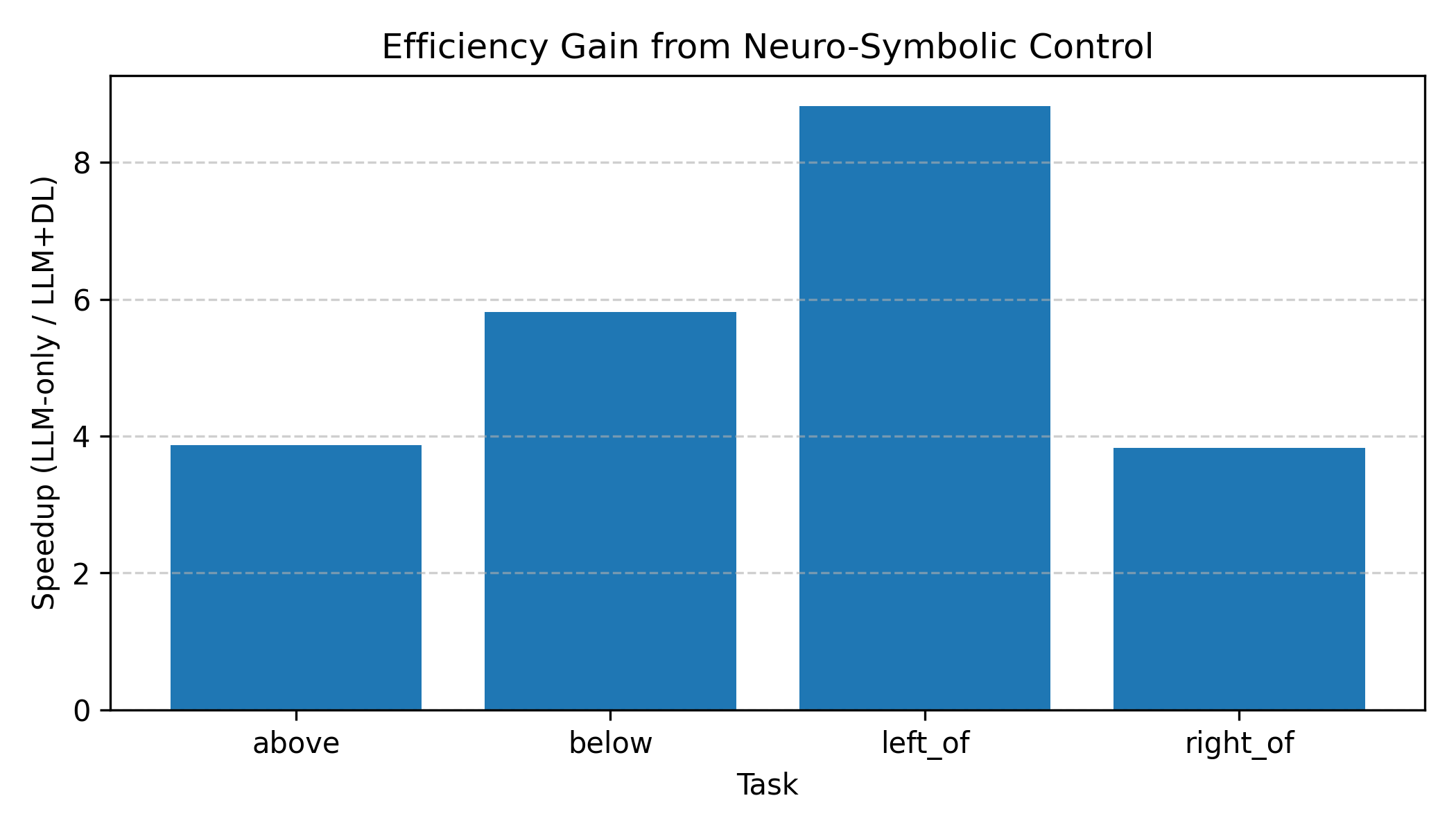}
    \caption{Speedup of LLM+DL relative to LLM-only control aggregated across language models. 
    The proposed framework achieves substantial efficiency gains, with speedups of up to 
    $8.83\times$ depending on the spatial relation.}
    \label{fig:speedup_task}
\end{figure}

For the significant speedup, for all tasks, the suggested method delivers an average speedup of $5.58\times$, with a maximum speedup of $8.83\times$ for the \texttt{left\_of} relation. For the task-dependent gains, compared to orthogonal relations (\texttt{right\_of}, \texttt{above}), lateral relations (\texttt{left\_of}, \texttt{below}) show greater increases in speedup and success rate ($\Delta\text{SR} = 0.48$). This imbalance implies that some spatial interactions are more difficult for LLM-only management because of incorrect or hallucinated coordinate predictions. Regardless of the kind of work, step reduction consistently surpasses 70\%, indicating that the neural delta controller significantly increases control efficiency.

As summarized in Figure \ref{fig:speedup_task}, the proposed method achieves speedups of up to $8.83\times$ over LLM-only control.


\begin{figure}[ht!]
    \centering
    \includegraphics[width=0.90\linewidth]{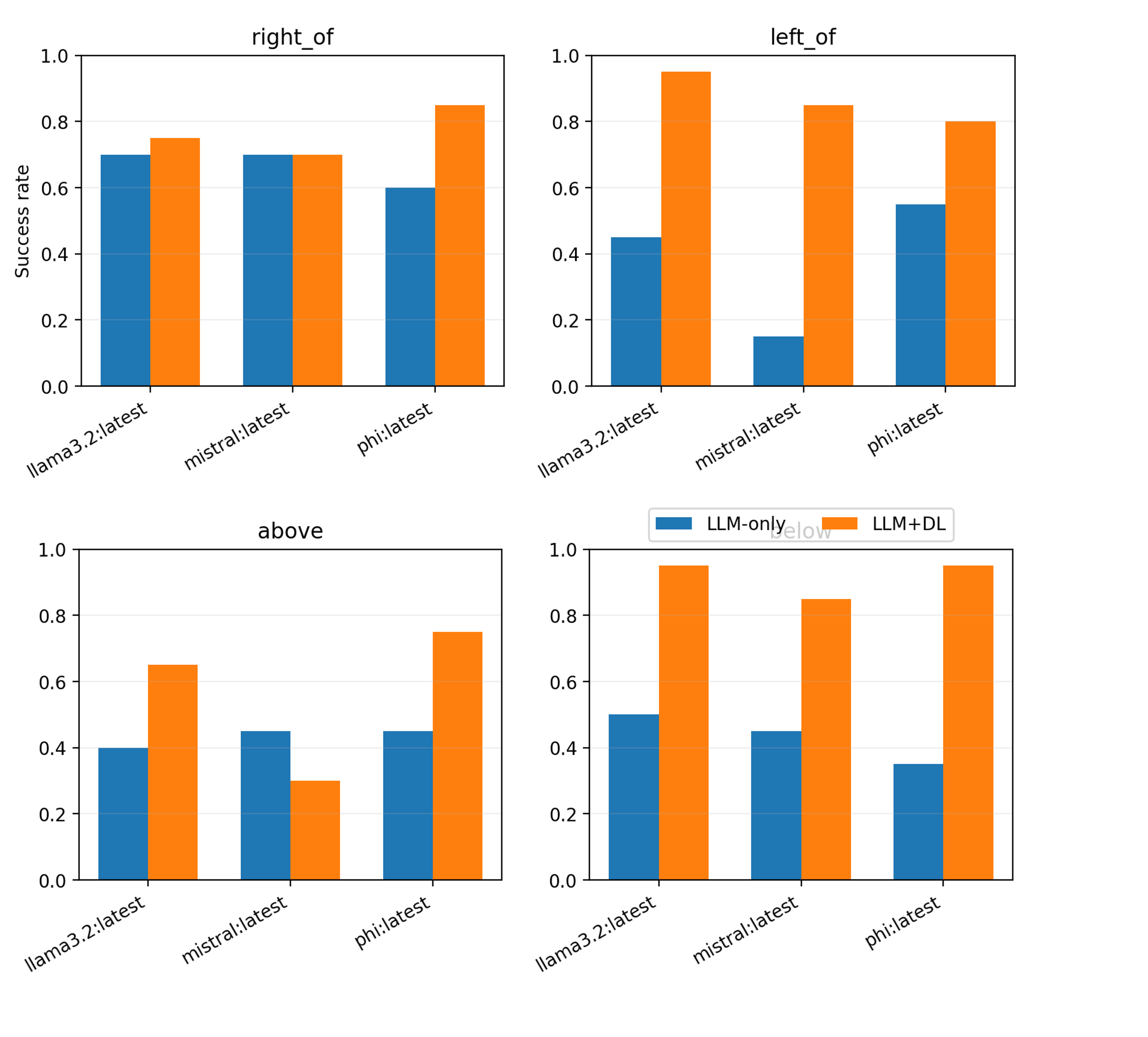}
    \caption{Success rate by language model and task. The suggested LLM+DL framework consistently improves performance on Mistral, Phi, and LLaMA-3.2, demonstrating resilience to scale and language model selection.}
    \label{fig:success_by_task_model}
\end{figure}

As the performance gains are consistent across with several notable observations:

\begin{itemize}
    \item For the model-agnostic improvement, the LLM+DL framework enhances all three models performance, suggesting that architectural decomposition, rather than language model capacity or scale, is the source of the advantages.
    
    \item For the compensation effect, even in situations where LLM-only control operates badly, robust performance is made possible by the neural controller's compensation for inferior symbolic thinking. This is especially true for Phi, where LLM+DL outperforms the model even with fewer parameters.
    
    \item Due to the decreased variance, the LLM+DL performs more consistently across episodes and beginning settings, as seen by a significantly lower standard deviation of success rates than LLM-only.
\end{itemize}
The performance improvements of the neuro-symbolic framework hold true for all assessed local LLMs, as shown in Fig. \ref{fig:success_by_task_model}.

\subsubsection{Ablation: Component Contribution}

We calculate the performance difference between approaches in order to further isolate each component's contribution:

\begin{align}
\Delta_{\text{symbolic}} &= \text{SR}_{\text{LLM+DL}} - \text{SR}_{\text{DL-only}}, \\
\Delta_{\text{neural}} &= \text{SR}_{\text{LLM+DL}} - \text{SR}_{\text{LLM-only}}.
\end{align}

Averaged across all configurations, we observe $\Delta_{\text{symbolic}} = 0.12$ and $\Delta_{\text{neural}} = 0.30$, indicating that both components contribute meaningfully to overall performance, with the neural controller providing the larger marginal improvement. We can infer the following inferences from the experimental results:

\begin{itemize}
    \item Control using LLM alone is ineffective: Particularly for spatial manipulation tasks requiring numerical precision, direct coordinate prediction by LLMs results in sluggish convergence, significant variance, and inconsistent behavior.
    
    \item DL-only control lacks semantic flexibility: While the neural controller converges rapidly with oracle task encoding, it cannot interpret natural language instructions or generalize to novel task formulations.
    
    \item Neuro-symbolic integration works well: By combining the complementing advantages of neural control and symbolic thinking, the suggested LLM+DL paradigm achieves the following:
    \begin{itemize}
        \item Up to $8.83\times$ speedup over LLM-only control
        \item Average success rate improvement of 0.30
        \item Consistent gains across all language models and tasks
        \item Reduced variance and improved robustness
    \end{itemize}
    
    \item Architectural design matters: The neuro-symbolic decomposition offers advantages independent of underlying model capability, as evidenced by the performance gains that are consistent across language models of different scales.
\end{itemize}

These results support the main hypothesis that more effective, reliable, and comprehensible behavior is produced in language-conditioned manipulation tasks when symbolic thinking and continuous control are separated.

\section{Discussion}
\label{sec:discussion}
\subsection{Deterministic Stability vs Motion Fluidity}
The symbolic discretization mainly prioritizes deterministic and stable execution by preventing
hallucinated actions, but it may produce less fluid trajectories in some cases. This trade-off
is intentional for the purpose of safety and convergence, and the framework can further be extended to finer
symbolic resolutions (e.g., headings or intensities) while preserving bounded control.

\subsection{Semantic Limits of Local Language Models}
Locally deployed language models may introduce a lower semantic ceiling than cloud-scale
options, specifically for abstract or compositional instructions. Since the neural
controller cannot correct incorrect symbolic intent, but it does stabilize execution once a
symbolic decision is made, and richer symbolic reasoning can be incorporated as per the need.

\subsection{Robustness to Sensor Noise and State Uncertainty}
The evaluation presented in the paper assumes ground-truth state information to isolate control behavior.
In the presence of moderate sensor noise, the closed-loop delta controller can modify and correct
transient symbolic errors as state estimates are updated, while persistent uncertainty
would require additional perceptual filtering.

\subsection{Generalization Beyond 2D Environments}
Although evaluated in a planar 2D setting, the neuro-symbolic decomposition presented in this paper is not
inherently restricted to only two dimensions. Extension to 3D environments would involve
increasing the state representation and symbolic predicate set. Irregular object
shapes can be accommodated through geometry-aware symbolic abstractions. This modular design
facilitates such extensions without changing the core execution controller.

\section{Conclusion}
\label{sec:discussion_conclusion}

In this paper a neuro-symbolic control framework for language-guided spatial
manipulation that explicitly separates high-level symbolic reasoning from low-level
continuous execution is presented. By constraining the language model to symbolic task outputs and
delegating numerical control to a neural delta controller, the proposed architecture
addresses key limitations of monolithic LLM-based control, including unstable trajectories,
slow convergence, and hallucinated actions. The experimental results demonstrate that this
decomposition leads to more stable and efficient closed-loop behavior.

Across all evaluated spatial tasks and locally deployed language models, the proposed
LLM+DL framework consistently outperforms LLM-only baselines in convergence efficiency
while preserving or improving task success rates. Average step reductions exceed 70\%,
with speedups of up to $8.83\times$, and these gains remain consistent even for smaller
language models with weaker symbolic reasoning capabilities. These findings indicate that
the observed performance improvements arise primarily from architectural design rather than
language model scale, highlighting the effectiveness of neuro-symbolic integration for
reliable and interpretable control.

Despite these advantages, the current study is limited to planar environments with perfect
state observability and a fixed set of primitive spatial relations. The expressiveness of the
symbolic layer is constrained by the predefined task vocabulary, and stochastic language
model outputs and inference latency remain practical considerations. Future work will
extend this framework to real-world robotic platforms, richer symbolic representations,
hierarchical and compositional task structures, and vision-based environments, with the
goal of enabling robust language-guided control in more complex and realistic settings.


\bibliographystyle{IEEEtran}

\newpage
\section*{Supplementary Material}
\label{sec:supplementary}

Additional experimental data and analysis that corroborate the conclusions in the main paper are provided in this supplementary section. These findings provide more clear and precise information about ablation results, task-wise convergence patterns, and model-specific behavior. Every experiment adheres to the same procedure outlined in Section~\ref{sec:experiments}.

\subsection*{S1. Success Rate by Language Model and Task}

\begin{figure}[ht!]
    \centering
    \includegraphics[width=0.90\linewidth]{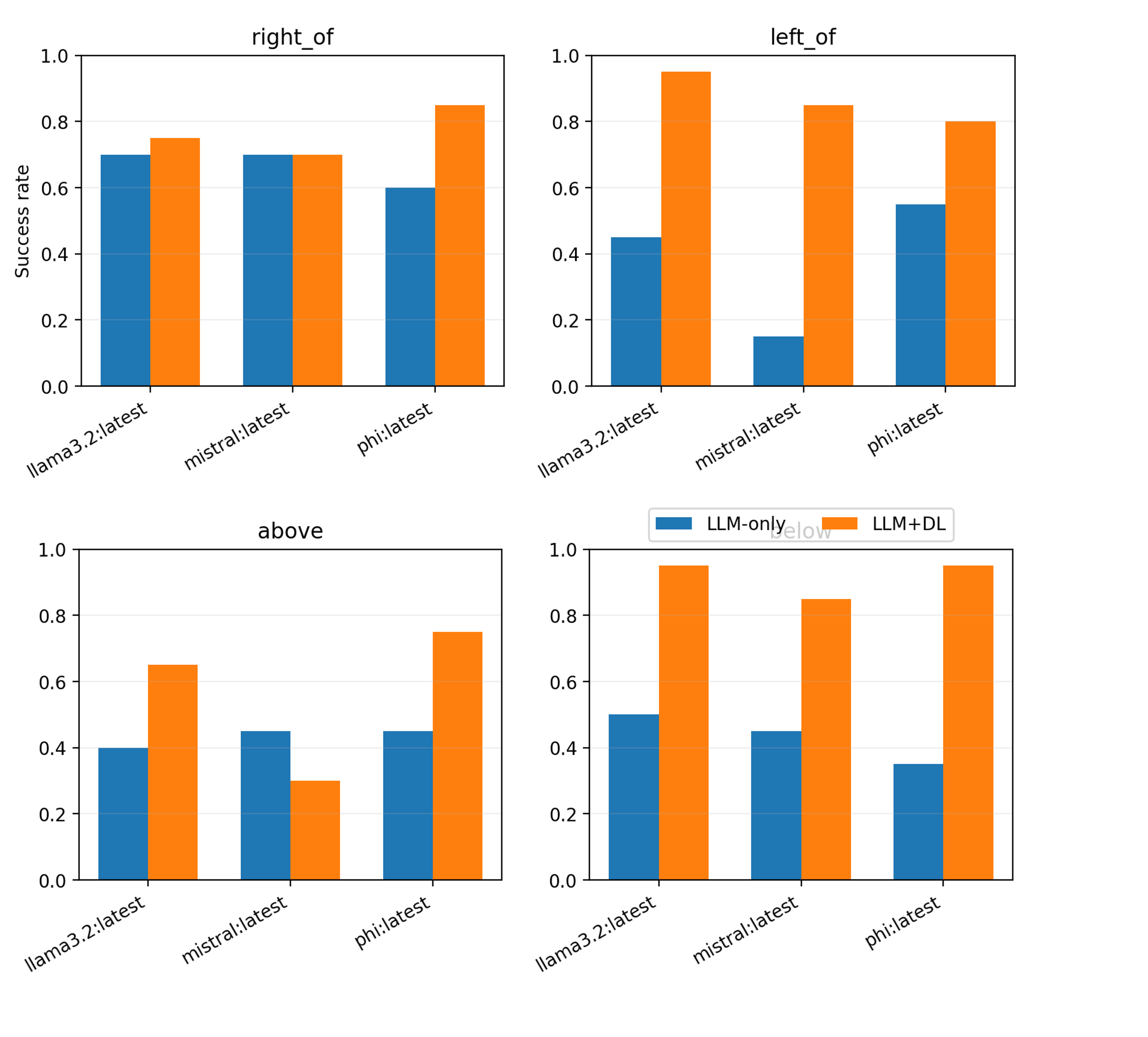}
    \caption{Success rate by task and language model. The neuro-symbolic LLM+DL framework improves reliability across all evaluated LLMs, including smaller models such as Phi.}
    \label{fig:supp_success_by_model}
\end{figure}

Success rates for each spatial challenge are displayed by language model in Figure~\ref{fig:supp_success_by_model}. The LLM+DL framework significantly increases success rates compared to LLM-only control in almost all configurations, while absolute performance differs among models.

\subsection*{S2. Average Steps by Language Model}

\begin{figure}[ht!]
    \centering
    \begin{minipage}{0.7\linewidth}
        \centering
        \includegraphics[width=\linewidth]{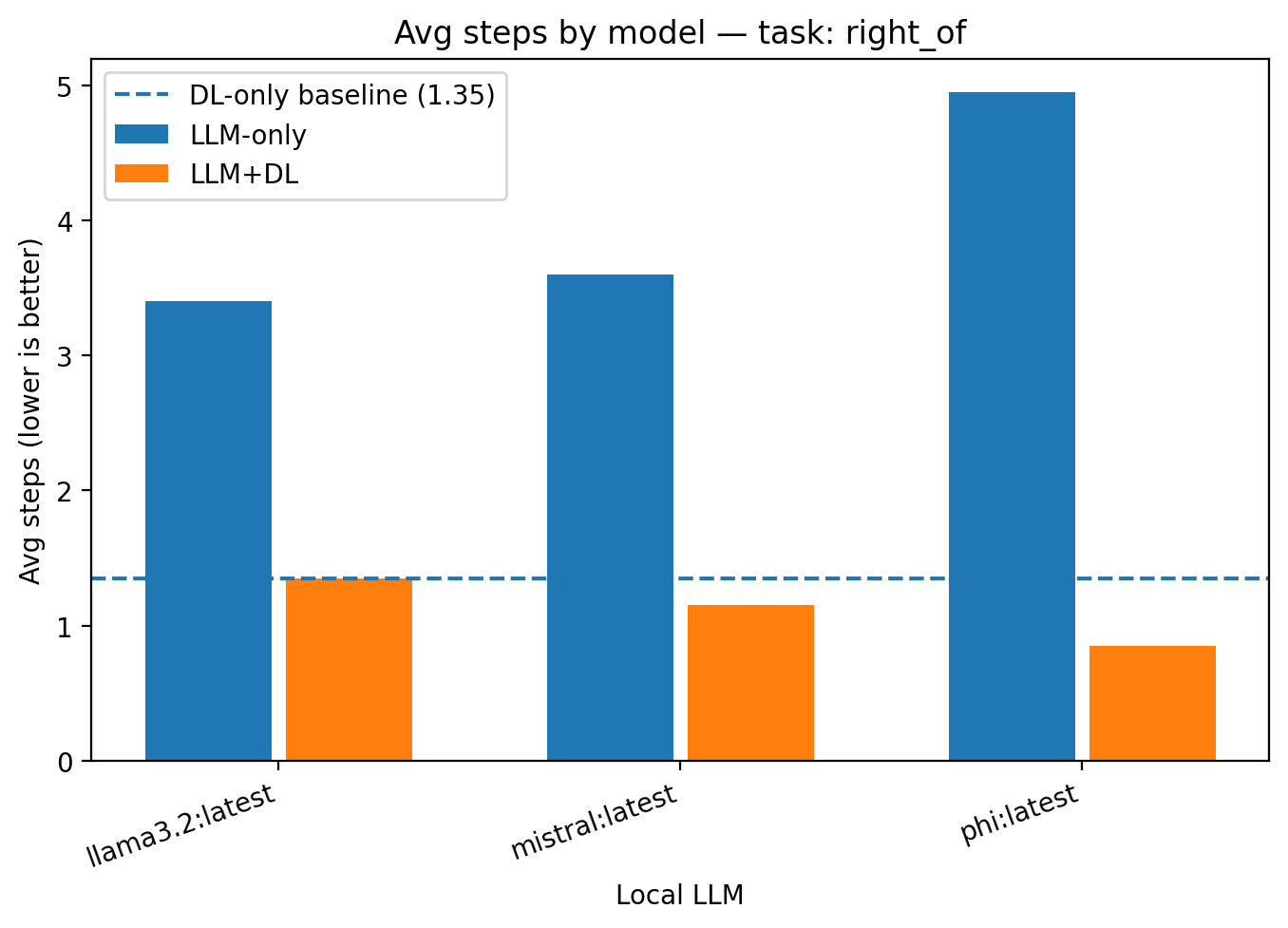}
    \end{minipage}
    \hfill
    \begin{minipage}{0.7\linewidth}
        \centering
        \includegraphics[width=\linewidth]{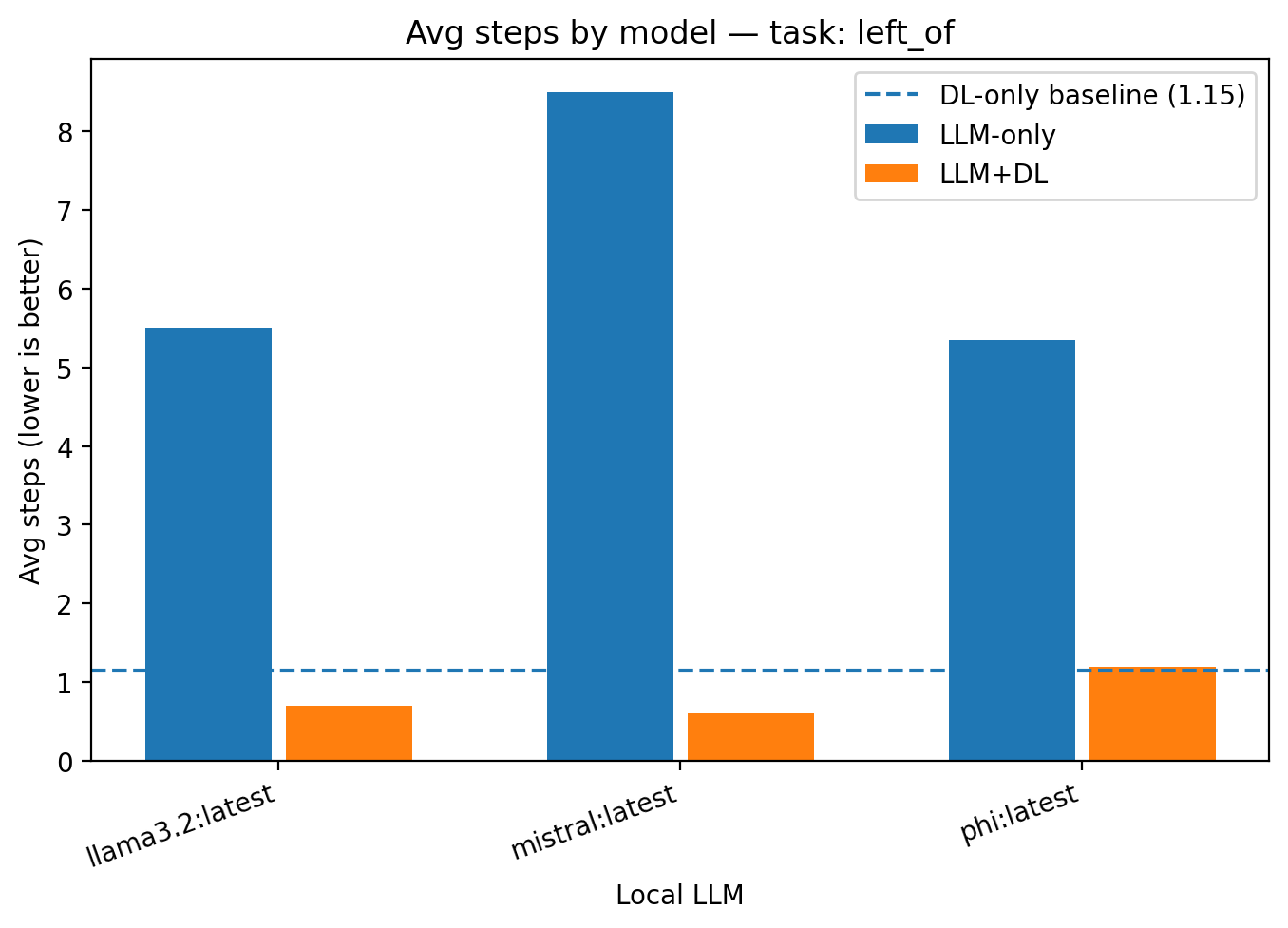}
    \end{minipage}

    \caption{The average number of control steps for the \texttt{left\_of} and \texttt{right\_of} tasks by language model. For all models, LLM+DL achieves significant step count reductions.}
    \label{fig:supp_avg_steps_model}
\end{figure}

Figure~\ref{fig:supp_avg_steps_model} provides an additional analysis of efficiency by breaking down the average number of steps needed for successful episodes by task and language model. While LLM+DL consistently converges faster across all models, the LLM-only control shows substantial volatility and noticeably increased step counts.

\subsection*{S3. Convergence Behavior Across Tasks}

\begin{figure}[ht!]
    \centering
    \begin{minipage}{0.48\linewidth}
        \centering
        \includegraphics[width=\linewidth]{convergence_norm_right_of.png}
    \end{minipage}
    \hfill
    \begin{minipage}{0.48\linewidth}
        \centering
        \includegraphics[width=\linewidth]{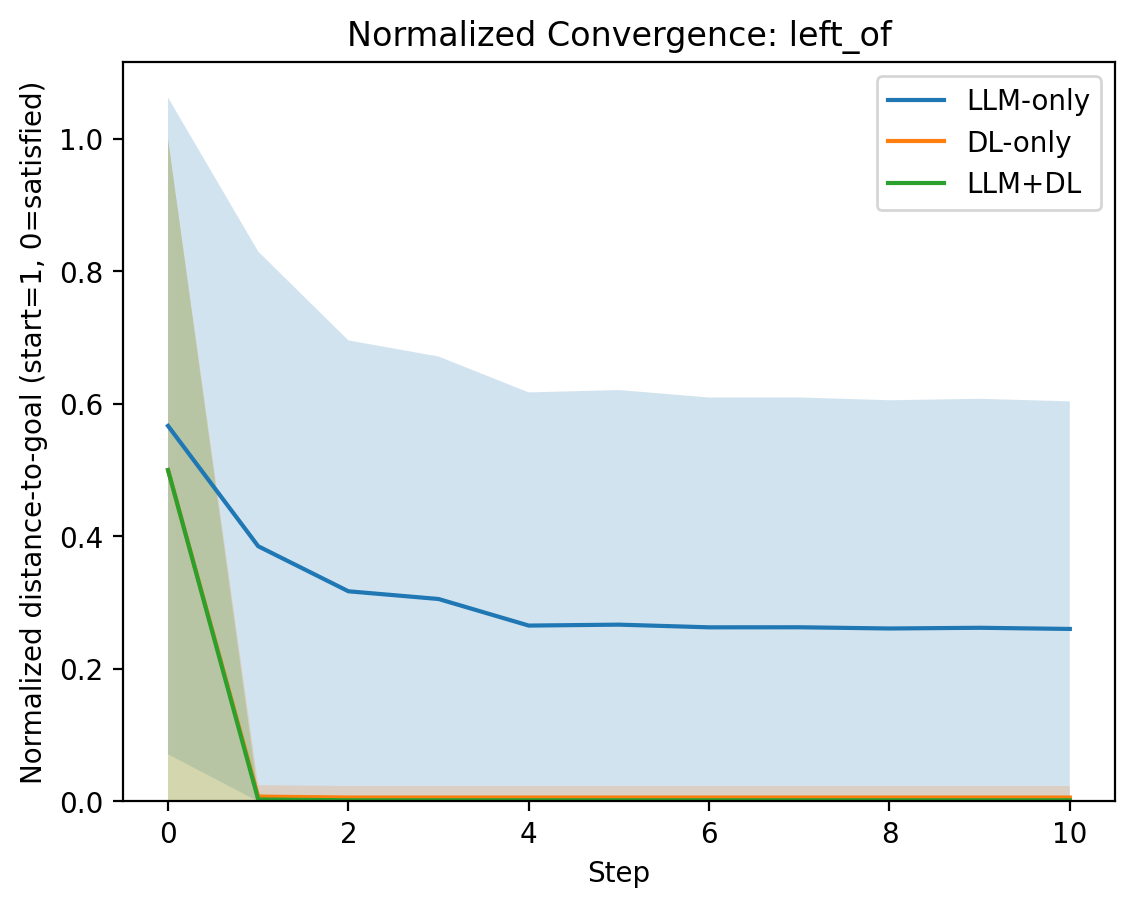}
    \end{minipage}

    \vspace{0.5em}

    \begin{minipage}{0.48\linewidth}
        \centering
        \includegraphics[width=\linewidth]{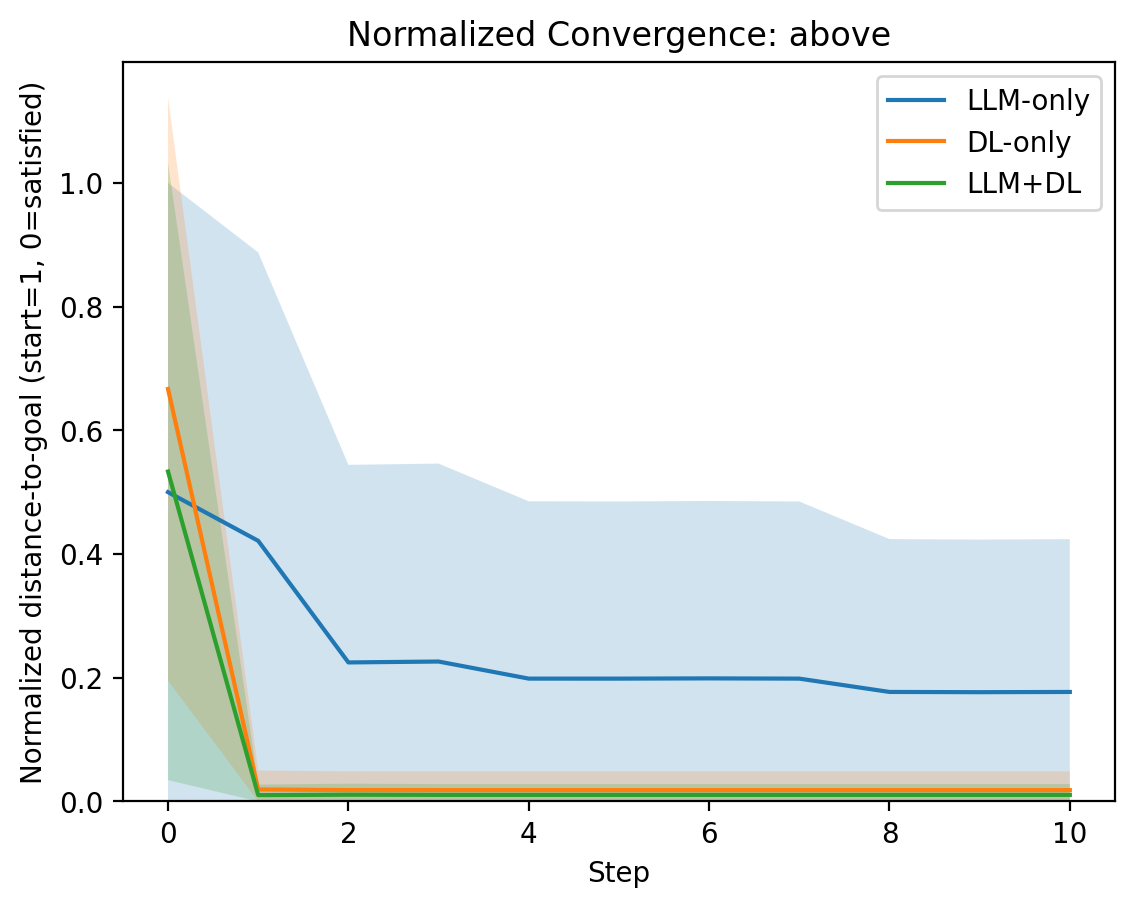}
    \end{minipage}
    \hfill
    \begin{minipage}{0.48\linewidth}
        \centering
        \includegraphics[width=\linewidth]{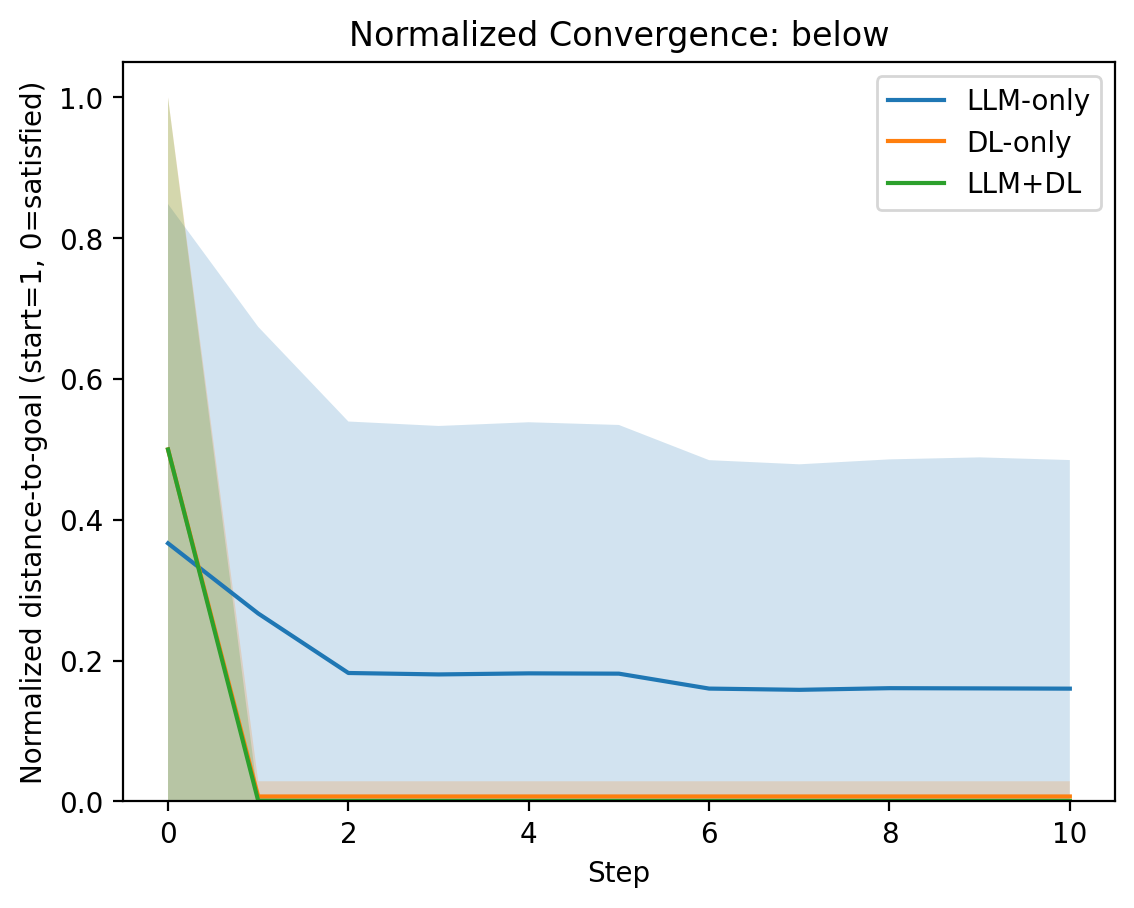}
    \end{minipage}

    \caption{Normalized distance-to-goal over time for all spatial tasks. The proposed LLM+DL framework converges faster and more stably than the LLM-only control across all relations.}
    \label{fig:supp_convergence_all}
\end{figure}

Normalized distance-to-goal curves for each of the four spatial relations are shown in Figure~\ref{fig:supp_convergence_all}. These figures show that while the LLM+DL framework generates smooth, monotonic trajectories with lower variance, LLM-only control frequently displays oscillatory or non-monotonic convergence.

\subsection*{S4. Model-Specific Relative Improvements}

\begin{figure}[ht!]
    \centering
    \begin{minipage}{0.65\linewidth}
        \centering
        \includegraphics[width=\linewidth]{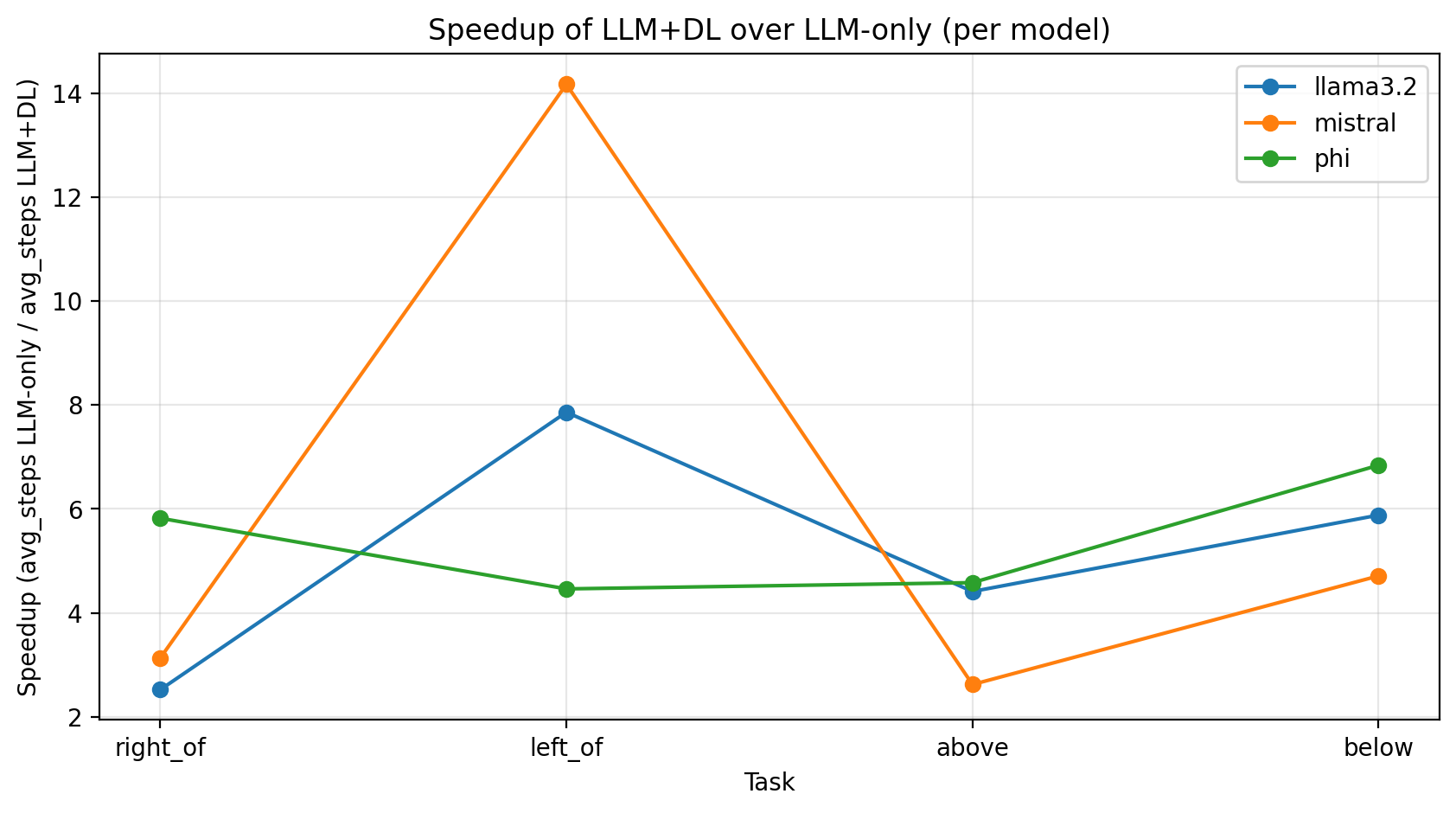}
    \end{minipage}
    \hfill
    \begin{minipage}{0.65\linewidth}
        \centering
        \includegraphics[width=\linewidth]{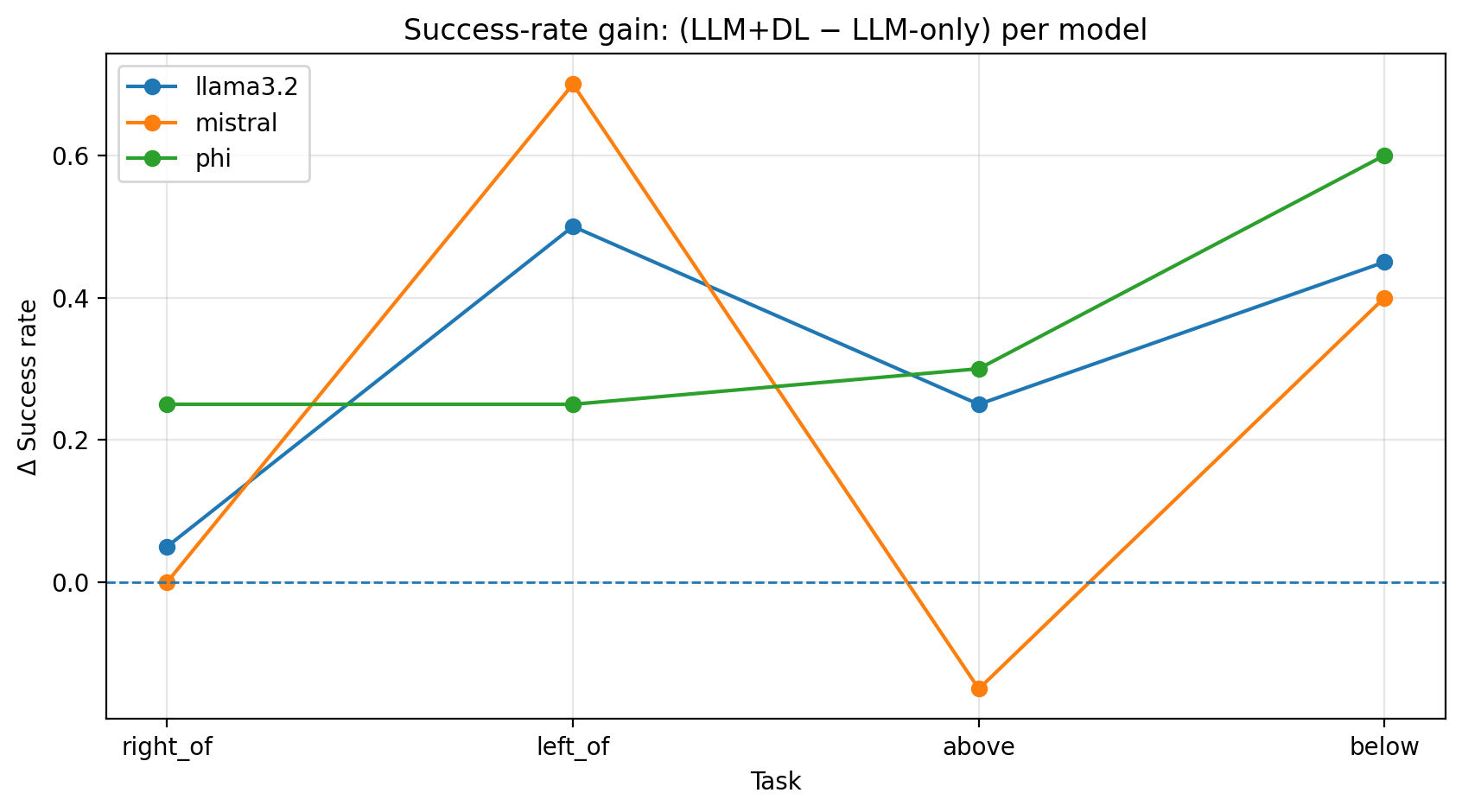}
    \end{minipage}

    \caption{Model-specific speedup (left) and success-rate improvement (right) of LLM+DL over LLM-only control. Performance gains are consistent across all evaluated language models.}
    \label{fig:supp_speedup_model}
\end{figure}

Figure~\ref{fig:supp_speedup_model} shows the speedup factors and success-rate increases of LLM+DL over LLM-only control for each language model in order to measure the advantage of neuro-symbolic integration at the model level.

\subsection*{S5. Discussion of Supplementary Results}

The additional findings support the paper's primary conclusions. In every in-depth analysis, the LLM+DL framework shows:

\begin{itemize}
    \item Gains in performance that are consistent across different-scale language models
    \item Considerable decreases in convergence time and control steps
    \item Reduced variation and increased stability in closed-loop behavior
\end{itemize}

The efficacy of distinguishing symbolic reasoning from continuous execution in language-conditioned control tasks is further supported by these results.

\end{document}